\documentclass[final,5p,times,twocolumn]{elsarticle}

\usepackage{amssymb}
\usepackage{amsmath,amsfonts}
\usepackage{algorithmic}
\usepackage{algorithm}
\usepackage{array}
\usepackage[caption=false,font=normalsize,labelfont=sf,textfont=sf]{subfig}
\usepackage{textcomp}
\usepackage{stfloats}
\usepackage{url}
\usepackage{verbatim}
\usepackage{cite}
\usepackage{amsmath}
\usepackage{amssymb}
\usepackage{booktabs}
\usepackage{multirow}
\usepackage{adjustbox}
\usepackage{xcolor}
\usepackage{graphicx}
\usepackage{xspace}
\usepackage{cleveref}
\usepackage{arydshln}
\usepackage[T1]{fontenc}

\makeatletter
\DeclareRobustCommand\onedot{\futurelet\@let@token\@onedot}
\def\@onedot{\ifx\@let@token.\else.\null\fi\xspace}
\makeatother

\def\eg{\emph{e.g.}} 
\def\ie{\emph{i.e}\onedot}

\def\etal{\emph{et al.}}

\newcommand{\loc}{\text{loc}}
\newcommand{\size}{\text{size}}

\newcommand{\ca}{\text{CA}}

\journal{ISPRS Journal of Photogrammetry and Remote Sensing}

\begin{document}

\begin{frontmatter}

%% Title, authors and addresses

%% use the tnoteref command within \title for footnotes;
%% use the tnotetext command for theassociated footnote;
%% use the fnref command within \author or \address for footnotes;
%% use the fntext command for theassociated footnote;
%% use the corref command within \author for corresponding author footnotes;
%% use the cortext command for theassociated footnote;
%% use the ead command for the email address,
%% and the form \ead[url] for the home page:
%% \title{Title\tnoteref{label1}}
%% \tnotetext[label1]{}
\author{Dimitri Gominski\corref{cor1}\fnref{label1}}
\ead{dg@ign.ku.dk}
\author{Ankit Kariryaa\fnref{label2}}
\author{Martin Brandt\fnref{label1}}
\author{Christian Igel\fnref{label2}}
\author{Sizhuo Li\fnref{label1}}
\author{Maurice Mugabowindekwe\fnref{label1}}
\author{Rasmus Fensholt\fnref{label1}}
%% \ead[url]{home page}
% \fntext[label1]{Department of Geosciences and Natural Resource Management, University of Copenhagen, Copenhagen, Denmark}
% \fntext[label2]{Department of Geosciences and Natural Resource Management, University of Copenhagen, Copenhagen, Denmark}
%% \cortext[cor1]{}
% \affiliation{organization={},
%             addressline={},
%             city={},
%             postcode={},
%             state={},
%             country={}}
%% \fntext[label3]{}

\title{Benchmarking Individual Tree Mapping with Sub-meter Imagery}

%% use optional labels to link authors explicitly to addresses:
% \author[label1,label2]{}
\affiliation[label1]{organization={Department of Geosciences and Natural Resource Management, University of Copenhagen},
            city={Copenhagen},
            country={Denmark}}
\affiliation[label2]{organization={Department of Computer
Science, University of Copenhagen},
            % addressline={},
            city={Copenhagen},
            % postcode={},
            % state={},
            country={Denmark}}

% \author{}

% \affiliation{organization={},%Department and Organization
%             addressline={}, 
%             city={},
%             postcode={}, 
%             state={},
%             country={}}

\begin{abstract}
There is a rising interest in mapping trees using satellite or aerial imagery, but there is no  standardized evaluation protocol for comparing and enhancing methods. In dense canopy areas, the high variability of tree sizes and their spatial proximity makes it arduous to define the quality of the predictions. Concurrently, object-centric approaches such as bounding box detection usually perform poorly on small and dense objects. It thus remains unclear what is the ideal framework for individual tree mapping, in regards to detection and segmentation approaches, convolutional neural networks and transformers. In this paper, we introduce an evaluation framework suited for individual tree mapping in any physical environment, with annotation costs and applicative goals in mind. We review and compare different approaches and deep architectures, and introduce a new method that we experimentally prove to be a good compromise between segmentation and detection.
\end{abstract}

%%Graphical abstract
% \begin{graphicalabstract}
% %\includegraphics{grabs}
% \end{graphicalabstract}

%%Research highlights
% \begin{highlights}
% \item Research highlight 1
% \item Research highlight 2
% \end{highlights}

\begin{keyword}
%% keywords here, in the form: keyword \sep keyword
Tree mapping \sep Object detection/segmentation \sep Benchmarking
%% PACS codes here, in the form: \PACS code \sep code

%% MSC codes here, in the form: \MSC code \sep code
%% or \MSC[2008] code \sep code (2000 is the default)

\end{keyword}

\end{frontmatter}

%% \linenumbers

%% main text
\section{Introduction}
\label{sec:intro}

Trees play a major role in ecosystems and in the global carbon cycle. Disturbances such as droughts, insects, fires and deforestation cause widespread losses in tree cover, creating the need for better tools to rapidly assess tree resources across large areas. Concurrently, there is a rising interest in assessing carbon sequestration (plantations) or release (deforestation, wildfires) to prepare adequate policies mitigating future global warming. In this context, remote sensing systems provide a valuable source of both scalable and precise information. World- \citep{santoro_detailed_2018} and region-wide \citep{brandt_unexpectedly_2020} estimates of tree cover, counts or above-ground biomass are available thanks to remote sensing-based approaches, but a universal approach for individual tree mapping has yet to be identified.

With deep learning and the growing availability of sub-meter resolution satellite imagery, there is undoubtedly an emerging potential for conducting tree-level mapping at large scale.
There are various ways to achieve individual tree mapping; \eg~by estimating and separating tree canopy cover (a paradigm commonly referred to as instance segmentation) or by delineating their extent with a box (object detection) among others. Segmentation requires a time-consuming contouring of instances and allows shape predictions, while box detection reduces objects to two spatial extents (plus possibly orientation \citep{ding_object_2021}). Although those approaches are typically used for different purposes in computer vision, they represent different ways of achieving the same goal within tree mapping from remote sensing imagery, namely inferring positions, sizes and other attributes of trees. Biomass for example can be estimated from crown diameter \citep{mugabowindekwe_nation-wide_2023}, \ie a single dimension which can be computed from a delineation (segmentation) or approximated with the extent of bounding boxes (box detection). Considering the high annotation cost of segmentation, it is worth studying to which extent approximating tree crowns with simpler shapes influence the accuracy of predicted crown diameters/area.

Different methods are usually associated with their dedicated evaluation framework and metrics. For example, a segmentation approach is evaluated by finding the common area between predictions and labels, and a detection approach is evaluated by counting realistic matches against false positives and negatives using a common area criterion. Metrics influence the directions in which methods are improved. If there are multiple ways of performing tree mapping, how can these be evaluated in a comparable way? A common evaluation framework is needed for translating the objectives of individual tree mapping into standardized and comparable metrics.

Trees represent challenging variations for mapping due to differences in size, shapes and context related to species and environmental conditions.. In forests, separating overstory trees with very little texture and color variation is a difficult task, even to the human eye. Methods must both predict a correct canopy coverage and localization of individual instances despite the continuous canopy cover. The assumption that trees crowns have a roughly circular shape is a useful prior in this regard, it opens the way for faster annotation and can help separating instances after segmenting tree cover. But to our knowledge, no deep learning-based tree detection approach makes use of such prior. 

Here, we explore how to formulate and evaluate the task of individual tree mapping from sub-meter resolution remotely sensed imagery, and propose solutions with deep learning-based tools. Our contributions are as follows: 

We propose a benchmark protocol to quantitatively measure detection, localization and characterization of individual trees. This evaluation lays the methodological tools for comparing methods toward large-scale mapping of trees.

We review and implement different deep learning architectures and frameworks to conduct this task.

Building on this comparison and the assumption of tree crown shapes being circular, we propose a new method for individual tree mapping. Our method combines the advantages of precise high-resolution segmentation approaches with the ability of detection approaches to separate instances, and does not need costly individual polygons.

We evaluate architectures and frameworks in a standardized way on an annotated dataset, and discuss  the experimental results to identify promising new research directions. Our proposed method shows good detection rates, low localization and size estimation errors on the most difficult scenario, and is particularly suited for model ensembling.

\section{Context}
\label{sec:context}

Depending on the application, products of woody resources provided as averaged values on low-resolution patches such as tree cover or biomass estimations at the stand level or species predominance might be sufficient, but represent in many cases a sub-optimal solution as compared to mapping trees at the level of individuals. Individual tree mapping opens the way for characterization of trees at individual level (crown size, height and species), supporting detailed management: resources management and carbon sequestration assessment, detection of logging, species distribution for biodiversity management, monitoring of restoration etc.; as well as precise inventories of trees outside forests, such as in urban areas.

\subsection{Related works}
\label{subsec:related_works}

Seen from satellites or airplanes, trees belong to the category of small objects with high intra-class variance. Detecting/segmenting objects in satellite or aerial imagery, \ie mapping objects, is a problem that has been thoroughly addressed with classes such as roads \citep{shamsolmoali_road_2021}, buildings \citep{sirko_continental-scale_2021}, vehicles \citep{koga_method_2020}. Remote sensing object mapping has immensely benefited from the advent of deep learning, taking advantage of models designed for "mainstream" computer vision such as R-CNN and its successors \citep{ren_faster_2015} and applying them to a broad range of objects with good results \citep{ding_object_2021}.

Objects in remote sensing images present unique challenges: they can be smaller than objects in common images due to the ground resolution of remote sensors, requiring tailored metrics \citep{xu_dot_2021} or models \citep{pang_r2-cnn_2019}. They can also have different appearances depending on the area or sensor \citep{xu_multi-level_2022}, and training data is expensive and laborious to produce due to the expertise needed for annotations.

The tree class is subject to all of these constraints. Additionally, trees have considerable intra-class variance due to different sizes and species, they can be found in many different environments across the planet, and they are particularly difficult to annotate in their natural clustered state: forests. These difficulties might explain why trees are hitherto absent from large-scale remote sensing objects datasets \citep{ding_object_2021, cheng_survey_2016}. Pioneering works have recently launched the race towards digital inventories of trees, but they differ in their approach and result formatting: Weinstein \etal \citep{weinstein_remote_2021} used multi-modal data and some handmade crown delineations to derive training data for a bounding-box detection model, while Brandt \etal \citep{brandt_unexpectedly_2020} used handmade crown delineations to produce tree cover predictions with a segmentation model for counting purposes. Even small scales studies do not seem to agree on a common way of mapping trees, with bounding box detection \citep{dersch_novel_2022}, counting \citep{chen_transformer_2022}, segmentation \citep{yang_tree_2009, freudenberg_individual_2022, g_braga_tree_2020, noauthor_nation-wide_2022} or heatmap peak detection approaches \citep{ventura_individual_2022} proposed but never compared. One has to note that only overstory trees can be identified from an aerial perspective. Li et al. \citep{li_deep_2023} have used field plot data from Danish forests to demonstrate that large trees can be mapped with a relatively low bias (13\%),  but the bias is high if considering the count of all woody plants (47\%).

\subsection{Scope}

Individual tree mapping involves distinguishing every single tree and inferring information about its properties (shape, species, biomass, etc). Contrarily to density based products, it produces a discrete list of trees, including as a minimum their location. This can be done with different sensors, such as terrestrial LIDAR \citep{liu_point-cloud_2021}, airborne LIDAR \citep{kalinicheva_multi-layer_2022}, airborne optical cameras \citep{onishi_explainable_2021}, satellite imagery \citep{brandt_unexpectedly_2020}. Despite the rich and informative
data produced from LIDAR deployed on a UAV (Unmanned Aerial Vehicle),  acquisitions from UAV has limited potential beyond localized studies
due to their limited range in spatial coverage and high technological
cost.

In this study, we use optical imagery at 20 cm ground resolution. Sub-meter resolution data can be sourced from some national airborne imagery campaigns or commercial high-resolution satellites. Here we use aerial imagery because it is high-quality data suited for benchmarking, but the results remain valid for satellite imagery where individual trees are visible, provided that the appropriate data processing tools have been applied (\eg cloud removal, pansharpening if necessary). We argue that it is essential to establish a solid basis for comparing methods at sub-meter resolution, before generalizing to lower resolution with satellite data. 

Annotators typically annotate the tree crown surface with a polygon, and subsequently utilize its area to estimate the crown diameter, a valuable indicator for e.g. biomass estimation \citep{jucker_allometric_2017}. Previous studies have shown a high correlation with field measured crown areas, but annotators tend to draw smaller crowns on the screen as compared to practitioners measuring trees in the field \citep{brandt_unexpectedly_2020}.  This study solely relies on aerial imagery and is therefore limited to trees visible from above the canopy cover, while the evaluation protocol is open to expansion for a broader definition of trees.

\section{Evaluating individual tree mapping}
\label{sec:evaluation}

\begin{figure*}
    \centering
    \includegraphics[width=0.9\linewidth]{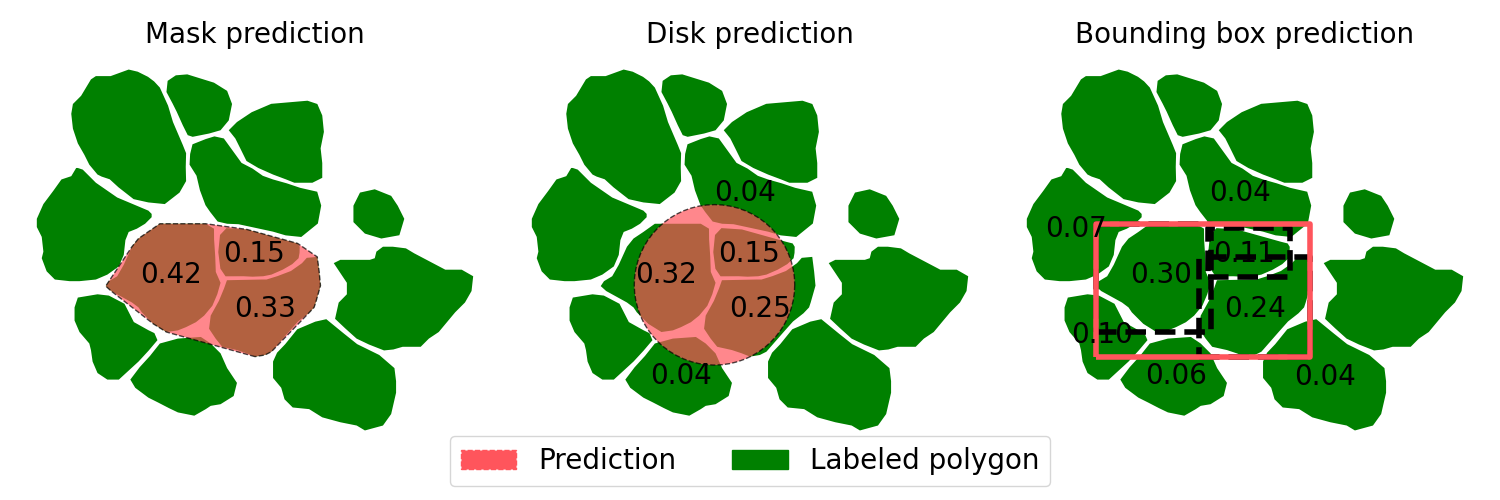}
    \caption{IoU measures with prediction forms, when predicting a large tree covering multiple labels.}
    \label{fig:iou_critic}
\end{figure*}

Individual tree mapping evaluation has to measure performance regarding three categories: are trees correctly detected - are trees correctly positioned - are tree attributes correctly estimated.

We focus here on the attribute that can be visually annotated from remote sensing imagery by a human expert (or field-measured \citep{li_deep_2023}), namely crown area. The same protocol can be used to evaluate tree height or above-ground biomass estimation, if additional information is available.

These categories correspond with the common errors made by models. In Section~\ref{sec:models} we will cover different ways of conducting tree mapping with deep models. Two broad families are heatmap-based models, where each each pixel is assigned a value related to tree presence, and anchor-based models, which output a list of detections with a center and extent. Accordingly, the errors from segmentation models come from wrongly labeled pixels, while detection models have a bilateral source of noise on locations and area values. Both families can produce false positives/negatives, shifted localizations or poorly estimated areas.

In denser areas, we argue that the main source of confusion for models is the proximity and large variability in size of tree crowns, leading to situations of a large prediction covering multiple trees or multiple small predictions within one tree. We verify experimentally in Section~\ref{sec:experiments} that this is the case both for labels and predictions. In this section, we describe an evalution framework specifically designed to handle these situations.

\subsection{Matching criterion}
\label{subsec:matching_crit}

On instance segmentation \citep{noauthor_coco_nodate, cordts_cityscapes_2016} or object detection \citep{everingham_pascal_2010} datasets, positives are typically defined with a threshold on IoU values between predicted and labeled masks or boxes, respectively. 

Consider a situation where a prediction covers three smaller labeled trees (see Figure~\ref{fig:iou_critic}). Regardless of how the prediction is produced (individual mask, center + area, bounding box), none of the labels gets an IoU above 0.5, the threshold commonly used to define positives when matching with IoU \citep{noauthor_coco_nodate}. A solution could be to set a lower threshold, but that would in practice include a significant number of neighboring trees if considering a model that predicts bounding boxes or disks, and produce poor matches in the case of mask prediction.

Instead, we consider the general case of a model outputing a tuple (position, area), and use a combination of distances between positions and crown areas. To associate predictions and labels, the list of $N$ labeled trees with center coordinates $p_i = (x_i, y_i)$ and crown area $\mathrm{CA}_i$ (and corresponding crown diameter $\mathrm{CD}_i$, $\mathrm{CA}_i = \pi (\frac{1}{2} \mathrm{CD}_i)^2$) is compared with the list of $M$ predictions with center coordinates $\hat p_j = (\hat{x_j}, \hat{y_j})$, estimated crown area $\hat{\mathrm{CA}_j}$. We define the cost of matching prediction $i$ with label $j$:
\begin{equation}
  c_{ij} = c^{\loc}_{ij} + \lambda c^{\size}_{ij}
  \label{eq:hungarian}
\end{equation}
with $\lambda$ a weighting factor. As the cost for size $c^{\size}_{ij}$, we use a simple L1 distance for symmetry. 

\begin{equation}
  c^{\size}_{ij} = \lVert \mathrm{CA}_i - \hat{\mathrm{CA}_j} \rVert_2 
  \label{eq:costsize}
\end{equation}

We enforce proximity between predictions and positives as a soft constraint through the localization cost $c^{\loc}_{ij}$. To ensure that matches remain spatially realistic, we add a hard constraint with the threshold $\gamma$ relative to the size of the labeled tree:

\begin{equation}
    \label{eq:cost}
    c^{\loc}_{ij}= \begin{cases} 
    \lVert p_i - \hat{p_j} \rVert_2 & \text{if} \lVert p_i - \hat{p_j} \rVert_2 < \gamma \mathrm{CD}_i \\
    \infty & \text{otherwise}
    \end{cases}
\end{equation}

$\gamma$ is the equivalent of the threshold in IoU-based matching.

We set $\lambda_{\size} = 0.1$ to have a balanced mean cost for the two criteria, in practice the weight can be adjusted depending on the applicative interest (for example, when evaluating methods with the goal of crown area estimation, $\lambda_{\size}$ can be set higher). In different set of experiments, we observed little variation in evaluation metrics with different $\lambda$ values, which indicates that the matching mostly depends on the localization cost (Eq. \ref{eq:cost}), the size criterion being merely a source of disambiguation in the case of clumped trees.

\subsection{Matching process}
\label{subsec:matching_proc}

\begin{figure}[H]
 \centering
   \includegraphics[width=0.96\linewidth]{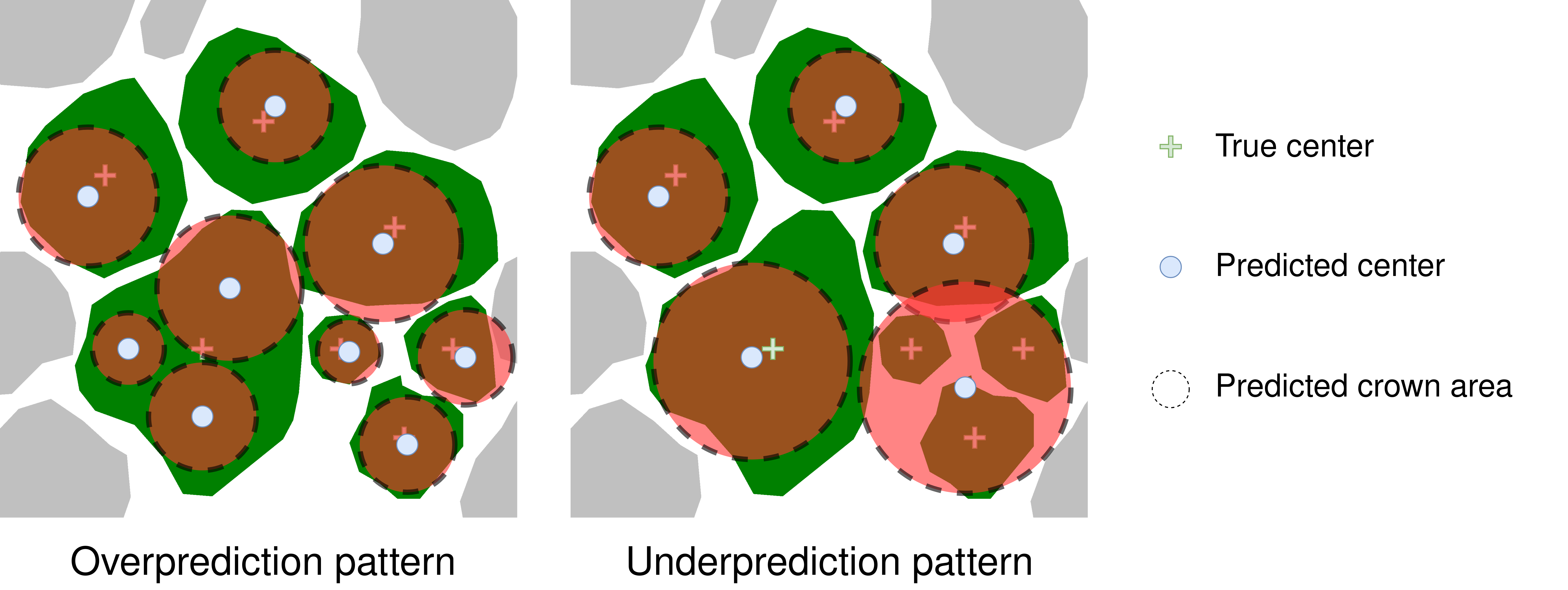}
   \caption{Examples of under- and overpredictions.}
   \label{fig:over_underpred}
\end{figure}

Tree detectors/segmentors can predict multiple small trees instead of a large one, and one large tree instead of multiple small ones (see Figure~\ref{fig:over_underpred}). While this is not the preferred outcome, it is better than having no prediction. The matching process associating predictions with labels for computing evaluation metrics can play a role in how those patterns are treated, by allowing a variable number of predictions or labels to be matched together. Here, we investigate how different matching processes influence detection metrics. 

We compute pairwise matching costs for $N$ labels and $M$ predictions in each patch, giving a $N$ x $M$ cost matrix $C$.

\textbf{One-to-one matching.} The Hungarian algorithm \citep{kuhn_hungarian_1955, song_rethinking_2021} produces a 1-to-1 assignment minimizing the total cost defined by $C$, outputting the matching matrix $A \in \mathbb{R}^{N\times M}$, with $A_{i,j} = 1$ if labeled tree $i$ is matched with prediction $j$, and $0$ otherwise.

\textbf{Many-to-one matching.} We match predictions with labels allowing multiple predictions to be associated to each label. This favors over-prediction. In practice, we do this by repeating the cost matrix $K$ times in the row direction, with $K$ being the maximum number of predictions matched with any label. This gives a $KN$ x $M$ cost matrix, after solving the assignment problem with the Hungarian algorithm we aggregate predictions per label and consider each label matched to at least one prediction as True Positive (TP).

\textbf{One-to-many matching}. We match predictions with labels allowing each prediction to be associated to multiple labels. This favors under-prediction. The threshold in equation~\ref{eq:cost} is replaced by $\gamma \hat{\mathrm{CD}_j}$ to account for large predictions covering small labeled trees. The matching is done by repeating the cost matrix $K$ times in the column direction, with $K$ being the maximum number of labels matched with any predictions.  This gives a $N$ x $KM$ cost matrix, after solving the assignment problem with the Hungarian algorithm we aggregate labels per prediction and consider each prediction matched to at least one label as True Positive (TP).

Labeled trees without a corresponding prediction are false negatives (FN) and predictions without a corresponding labeled tree are false positives (FP). The F1 score can be computed as $2\mathrm{TP} / (2\mathrm{TP} + \mathrm{FP} + \mathrm{FN})$.

\begin{figure}[H]
 \centering
   \includegraphics[width=0.96\linewidth]{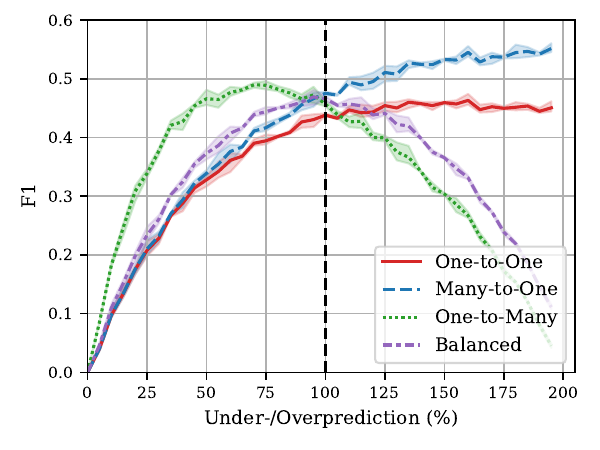}
   \caption{F1 score with under-/overprediction errors, by matching process.}
   \label{fig:f1_by_matching}
\end{figure}

We simulate an under- or over-predicting model with bigger and smaller predictions respectively (Figure~\ref{fig:f1_by_matching}). We randomly sample $sN$ labels $(p, \mathrm{CA})$, scale their crown area $\hat{\mathrm{CA}} = \text{min}((2 - s)\mathrm{CA}, 0)$, and shift the centers $\hat{p} \sim N(p, \mathrm{CD}/2)$. We then match the noisy generated predictions with the original labels with the different matching schemes, and measure the F1 score. One-to-many matching naturally gives a higher F1 score when underpredicting large trees (smaller sampling percentage $s$), while many-to-one matching gives a higher F1 when overpredicting small trees. One-to-one matching slightly favors overprediction of small trees, and gives a noticeably slow decrease of F1 score with more predictions. In contrast, one-to-many matching quickly drops to low F1 values due to the hard threshold being linked to predicted size.

To make use of the flexibility offered by one-to-many/many-to-one matching, we compute a composite metric, the balanced F1 score (bF1):

\begin{equation}
    \mathrm{bF1} = \alpha \mathrm{F1^{MO}} + (1 - \alpha) \mathrm{F1^{OM}}
    \label{eq:balanced_f1}
\end{equation}
\begin{equation}
    \alpha = \frac{1}{1 + e^{2\epsilon}}
    \label{eq:f1_sigmoid}
\end{equation} 
\begin{equation}
    \epsilon = \frac{M - N}{N}
    \label{eq:epsilon}
\end{equation}

with $\mathrm{F1^{MO}}$ the F1 score after many-to-one matching and $\mathrm{F1^{OM}}$ the F1 score after one-to-many matching.

$\alpha$ tunes the importance given to each metric depending on the normalized counting error: when overpredicting, $\epsilon >> 0 \Rightarrow \alpha \to 0$ therefore a higher weight for one-to-many matching ; when underpredicting, $\epsilon << 0 \Rightarrow \alpha \to 1$ therefore a higher weight for many-to-one matching. In other words, the balanced F1 score allows one-to-many and many-to-one matching, but diminishes their influence when they are computed in their preferential prediction regime. Figure~\ref{fig:f1_by_matching} shows the behavior of the composite metric. It reaches it maximum at $\epsilon \simeq 0$, is close to symmetric in the 0-2 range and converges to 0 in both directions. \\\\

\subsection{Metrics}
\label{subsec:metrics}

In addition to the balanced F1 score and normalized couting error $\epsilon$, we measure tree localization accuracy and crow area estimation accuracy.

Tree localization accuracy is measured with the Euclidean distance between a given labeled tree location and the associated prediction $\lVert p_i - \hat{p_j} \rVert_2$. Similarly to the balanced F1 score, we introduce the balanced localization error. With many-to-one matching, multiple matched predicted centers are first aggregated by averaging their positions before computing the $l_2$ distance. With one-to-many matching, multiple matched labeled centers are averaged.

\begin{multline}
    \text{Balanced } E_{\loc} = \frac{\alpha}{M} \sum_{i=0}^{N} \left(\lVert p_i - \frac{1}{M} \sum_{j=0}^{M} A^{MO}_{i,j} \hat{p_j} \rVert_2\right) \\
   + \frac{1 - \alpha}{N}  \sum_{j=0}^{M} \left(\lVert \hat{p_j} - \frac{1}{N} \sum_{i=0}^{N} A^{OM}_{i,j} p_i \rVert_2\right)
\end{multline}
where $A^{MO}$ is the aggregated matching matrix after many-to-one matching, with $A_{i,j}^{MO} = 1$ if and only if label $i$ is matched with prediction $j$ ; and reciprocally $A_{i,j}^{OM}$ is the aggregated matching matrix after one-to-many matching. By using the average location we make sure multiple predictions/labels align well overall with the matched tree.

We also report the balanced crown area estimation accuracy:
\begin{multline}
    \text{Balanced } E_{CA} = \frac{\alpha}{M} \sum_{i=0}^{N} \left(\lVert CA_i - \sum_{j=0}^{M} A^{MO}_{i,j} \hat{CA_j} \rVert_2\right) \\
   + \frac{1 - \alpha}{N}  \sum_{j=0}^{M} \left(\lVert \hat{CA_j} - \sum_{i=0}^{N} A^{OM}_{i,j} CA_i \rVert_2\right) 
\end{multline}
We aggregate multiple crown areas with sum, to take into account total crown area prediction. This aligns well with downstream tasks of tree cover or biomass mapping.

\subsection{Label noise}
\label{subsec:labnoise}

In reality, labels can also be subject to merging and splitting effects compared to real trees. If $\mathcal{T}$ is the set of real trees, $\mathcal{L}$ the set of labeled trees, and $\mathcal{P}$ the set of predicted trees, $\mathcal{L}$ is already corrupted with labeling errors, some labels correspond to 1/Nth of a tree in $\mathcal{T}$, and other labels correspond to N trees in $\mathcal{T}$. For simplicity, we assume that there is no situation of N trees in $\mathcal{L}$ covering M trees in $\mathcal{T}$, we verify in subsection \ref{subsec:labelexp} if this is realistic. We also ignore labels and predictions corresponding to no real trees and vice-versa in the following, without loss of generality.

We represent trees in $\mathcal{L}$ by nodes containing a quantity of real trees $Q_l$, with $Q_l \in \{\frac{1}{n} | n \in \mathbb{N} \} \cup \mathbb{N}$. We model the probability of having quantity $q$ in each node by a mixture of a one-truncated Poisson distribution, an inverse one-truncated Poisson distribution, and a Dirac distribution at 1:

\begin{equation}
    p(Q_l = q) = \begin{cases}
        p_{1}^{l} & \text{if } q = 1 \\
        \frac{1 - p_{1}^{l}}{2} \cdot \frac{f_p(q ; \lambda)}{1 - f_p(0 ; \lambda) - f_p(1 ; \lambda)} & \text{if } q = 2, 3, 4, \ldots \\
        \frac{1 - p_{1}^{l}}{2} \cdot \frac{f_p(\frac{1}{q}; \lambda)}{1 - f_p(0 ; \lambda) - f_p(1 ; \lambda)} & \text{if } q = \frac{1}{2}, \frac{1}{3}, \frac{1}{4}, \ldots \\
    \end{cases}
    \label{eq:labelquantity}
\end{equation}
where $\mathrm{f_p}$ is the standard Poisson pdf. See \ref{appx:labelnoisemodel} for a visual representation.

When we predict $\mathcal{P}$ with a tree detector and compare with $\mathcal{L}$, we again have possible splitting and merging. The real quantities in $\mathcal{P}$ will depend both on the quantities in $\mathcal{L}$ and on the quality of the predictions. We model this by associating each label node to a random number of predictions that depends on the label node quantity. A number above 1 simulates multiple labeled trees being matched to one prediction (labels are duplicated with equal quantities), while a number below 1 simulates multiple predicted trees being matched to one label. Again, we use a mixture of distributions, with a mixture weight that depends on the label quantity. We make the assumption that large labeled trees will have a higher chance of being predicted as multiple small trees, and small labeled trees will have a higher chance of being predicted as a single large tree.

\begin{equation}
    p(N = n | Q_l = q) = \begin{cases}
        p_{1}^{p} & \text{if } n = 1 \\
        q \frac{1 - p_{1}^{p}}{1 + q} \cdot \frac{f_p(n ; \lambda)}{1 - f_p(0 ; \lambda) - f_p(1 ; \lambda)} & \text{if } n = 2, 3, 4, \ldots \\
        \frac{1 - p_{1}^{p}}{1 + q} \cdot \frac{f_p(\frac{1}{n} ; \lambda)}{1 - f_p(0 ; \lambda) - f_p(1 ; \lambda)} & \text{if } n = \frac{1}{2}, \frac{1}{3}, \frac{1}{4}, \ldots \\
    \end{cases}
\end{equation}

The matching process influences what is considered positive and negative. With one-to-one matching, one prediction and one label at most are considered positive for each pair of prediction and label nodes. With many-to-one matching, one label at most is positive, with one-to-many matching, one prediction at most. We can also report what the optimal many-to-many matching would give by considering all connected predictions and labels positive. We measure precision and recall, here comparing to the real trees rather than the labeled ones.

\begin{figure}[t]
  \centering
   \includegraphics[width=\linewidth, trim={0 0.9cm 0 0}, clip]{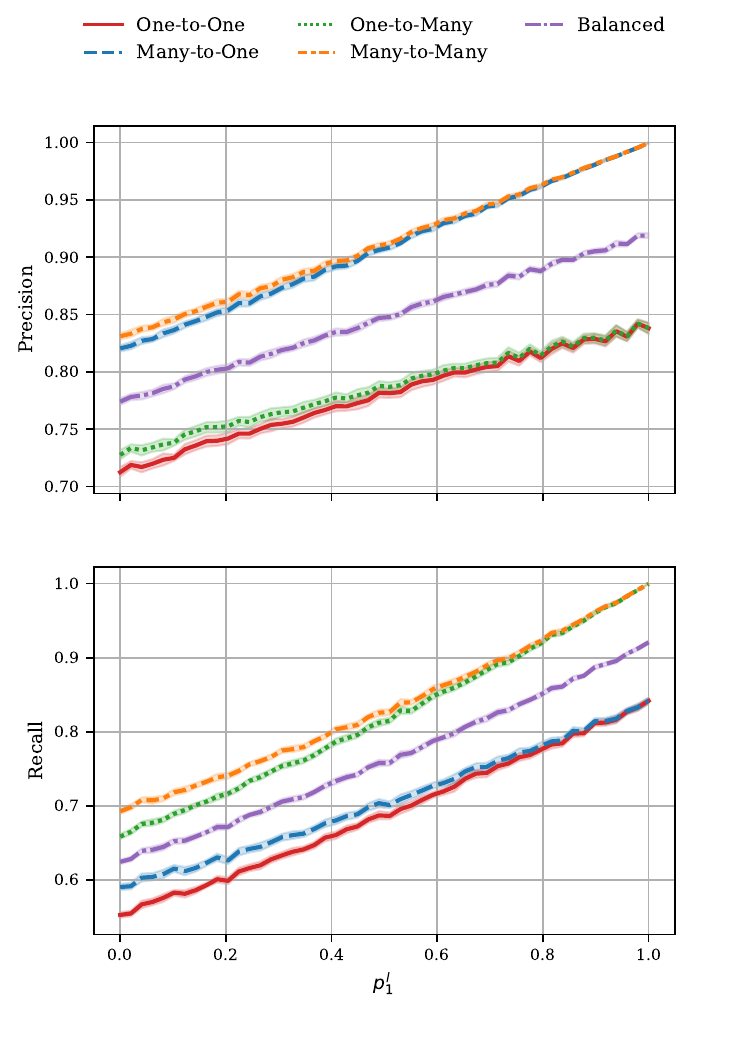}
   \caption{Precision and recall depending on label noise. Low $p_{1}^{l}$ means that real trees are likely to be merged or splitted, high $p_{1}^{l}$ means that real trees are likely to be correctly labeled as one tree.}
   \label{fig:label_noise}
\end{figure}

We simulate an imprecise annotator in Figure~\ref{fig:label_noise}, \ie an annotator that has a tendency of merging/splitting real trees. Many-to-many matching naturally leads to the best precision and recall, because it can fully capture the real distribution of trees. One-to-many matching and many-to-one matching give a tradeoff situation where either precision or recall is prioritized. One-to-one matching is the most conservative approach, with low precision and recall.

We simulate a biased annotator in Figure~\ref{fig:label_noise}, \ie an annotator that has a tendency towards annotating large trees covering multiple smaller real trees (higher second term in Eq~\ref{eq:labelquantity}) or conversely, towards annotating multiple small trees covering one larger real tree (higher third term in Eq~\ref{eq:labelquantity}). With a model kept equal, precision favors under-labeling (predictions are more often correct when labels merge real trees into bigger trees), while recall favors over-labeling (it is easier to recover real trees when labels split them into smaller trees). Time constraints probably push annotators to underlabel rather than overlabel. The same conclusions are reached here, with one-to-one matching being the most conservative and many-to-many matching being the best fit. 

Overall, one-to-one matching has low precision and recall, regardless of the annotation quality. This means that evaluation protocols with one-to-one matching will inevitably lead to an underestimation of how well methods can reproduce real trees. In contrast, many-to-many matching offers the possibility of being the closest to real performance, but is unpractical to implement. It can also hide severe effects of splitting or merging, since it does not take into account over/under-prediction tendency.

Our proposed balanced F1 score constitutes a middle ground between many-to-one matching and one-to-many matching. The trade-off between precision and recall is automatically set by $\alpha$, but could also be manually set. Values close to 1 prioritize high precision, low recall (emphasis on single tree detection or tracking over time), while values close to 0 prioritize low precision, high recall (emphasis on aggregated metrics such as biomass or counts per area).

\begin{figure}[t]
  \centering
   \includegraphics[width=\linewidth, trim={0 0.9cm 0 0}, clip]{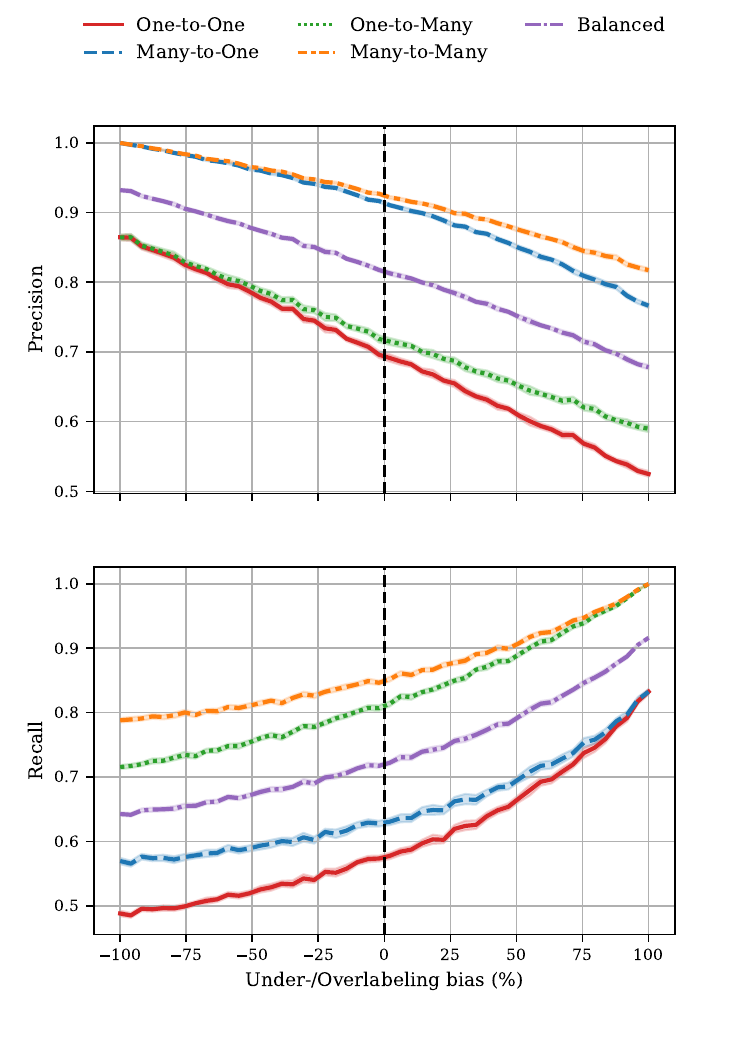}
   \caption{Precision and recall depending on label bias. Bias < 0 means that the annotator has a tendency of under-labeling large trees, bias > 0 means that the annotator has a tendency of over-labeling small trees.}
   \label{fig:label_bias}
\end{figure}

\section{Models and frameworks}
\label{sec:models}

\begin{figure*}[t]
  \centering
   \includegraphics[width=\linewidth]{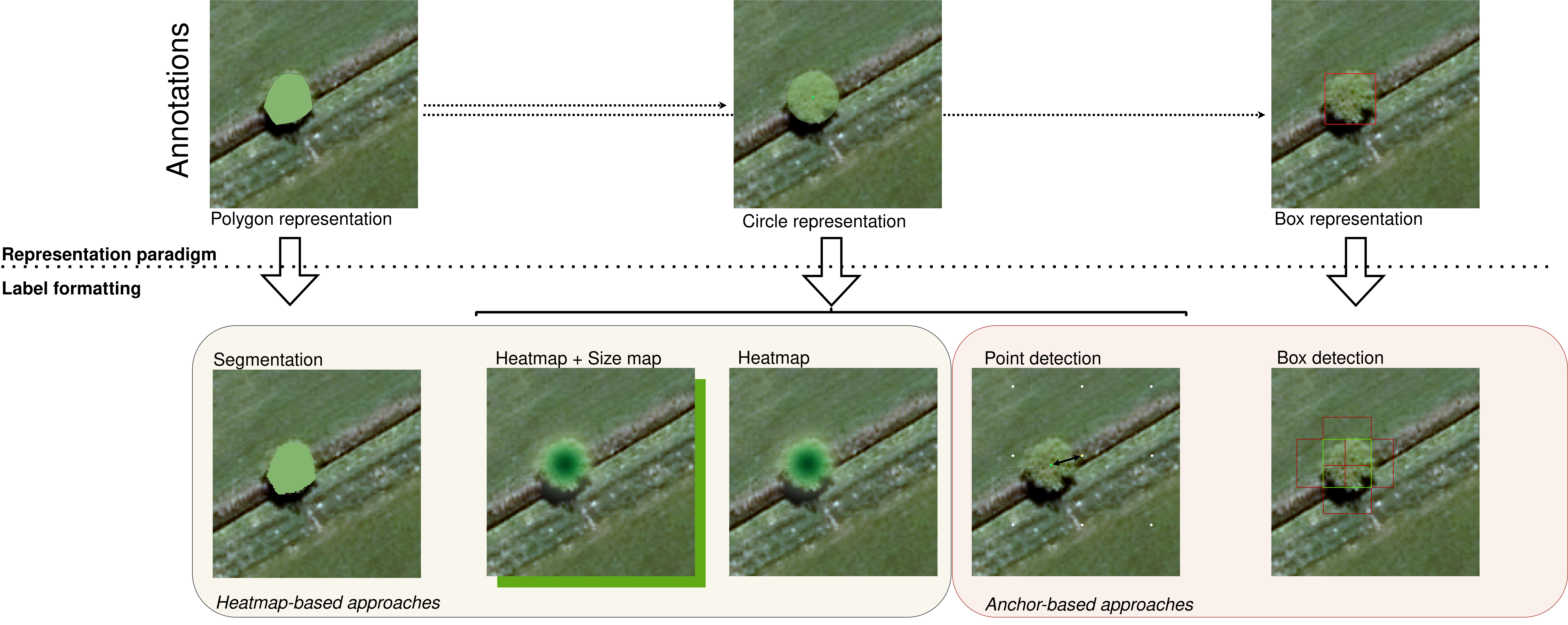}

   \caption{Different ways of performing tree mapping: methods differ by how they inherently treat objects (top row), and by how they format the labels to conduct the optimization process (bottom row). Heatmap-based approaches process the labels as a grid where each pixel is assigned a probability of tree presence or tree center, whereas anchor-based approaches rely on a fixed set of anchors that are matched to actual objects for training. }
   \label{fig:approaches}
\end{figure*}

Mapping can be understood as detection or instance segmentation in computer vision, but there are important differences to how these tasks are formulated in general-purpose computer vision. Typical datasets such as ImageNet or COCO are characterized by having a high number of classes, with a low number of instances per images and relatively big objects. These specificities have influenced modern segmentation and detection models, for example detection models are notoriously less accurate on small objects \citep{cai_bigdetection_2022}, and segmentation models have a tendency to blur object boundaries \citep{cheng_boundary_2021}.

Tree mapping is characterized by a high number of small objects, with repetitive positive examples in local areas but extreme variations in positive (tree shapes, species, sizes) and negative (vegetation, topography) patterns that can cause confusion. Similar tasks include medical image segmentation (large images, confusing inter-class similarities), crowd counting/localization (very small and repetitive objects), aerial object detection. The end goal is similar (locations, classes and attributes of objects), but the ways of treating the problem and the corresponding annotations vary. 

To provide a complete benchmark, we will consider different frameworks, based on how they treat and use the labeled data to produce predictions. We start by reviewing how the problem can be formulated, then compare architectural choices for deep feature extraction.

\subsection{Frameworks}
\label{subsec:frameworks}

We can consider multiple ways of performing individual tree mapping (Figure~\ref{fig:approaches}):

Pixel classification, or segmentation, is conducted by outputting a map with the same dimensions as the input image (or downscaled), where each pixel is given a binary value indicating tree presence by thresholding an intermediate heatmap. After inference, instances need to be separated to identify centers and sizes of trees. This can be done with morphological operations or energy-based assignment. We refer to this framework as \textit{Segmentation}.

Heatmap-based detection relies on an intermediate map giving for each pixel the confidence of object presence. The convolution operation in modern deep feature extractors encourages correlated values on local patches, and is thus much more efficient with smooth heatmaps as compared to sparse, binary values indicating object centers. A classical approach is therefore to preprocess the object centers with Gaussian kernels to produce a target heatmap. A recent example of this idea is CenterNet \citep{zhou_objects_2019}, where the model outputs a class-specific heatmap of "objectness", associated with an offset map (correcting the small deviations due to the limited gridsize) and a size map. By simply identifying peaks in the heatmap and getting the offset and size values at the corresponding locations, object proposals are obtained in a single forward pass. We propose to implement this idea in the context of tree mapping, by replacing the 2-dimensional size map for bounding boxes (width, height) by a 1-dimensional size map for circles (radius), and using a high-resolution backbone to have good localization accuracy without the offset map. This implies representing trees as disks rather than polygons, but considering that polygon-based methods reduce the area information to a single crown diameter value for downstream applications, this simplification is relevant. We refer to this framework as \textit{CenterNet}.

Detection for common objects is usually treated through the box representation: each object is associated with a bounding box that jointly expresses its location and spatial extent. The main challenge with this approach is the variable number of objects per image. A common solution is to use a fixed number $M$ of proposals with a grid of potential locations and object sizes, the anchors. Popular methods associate labeled objects with the predictions and use heuristic sampling to avoid computing the loss function on too many negative examples \citep{liu_ssd_2016, girshick_rich_2014}. We implement the well-known FasterRCNN \citep{ren_faster_2015} model by producing bounding boxes from the annotated polygons. During inference, the crown diameter is obtained as the mean of width and height of the predicted box. We refer to this framework as \textit{Box detection}.

The bounding box representation has its limits, especially in the context of tree mapping in view of their circular shape. Some works have departed from this approach by using keypoints instead, notably corners \citep{law_cornernet_2020} or centers \citep{song_rethinking_2021}. The latter is particularly relevant in our case, considering that we can represent trees as disks. We thus implement the P2P \citep{song_rethinking_2021} framework, which has the additional benefit of using the Hungarian algorithm to associate labeled points to predictions for computing the loss, thereby opening the way to direct optimization of our proposed evaluation metrics. This anchor-based approach is however limited to a fixed number of anchors, that must be kept relatively low to limit computational cost and allow proper optimization. For each point, the model outputs a class score, a positional offset to refine the predicted location, and a size prediction. We refer to this framework as \textit{Point detection}. 

\subsection{Heatmap detection}
\label{subsec:heatmapdet}

\begin{figure}[!t]
\centering
\hfil
\subfloat[]{\includegraphics[width=.85in]{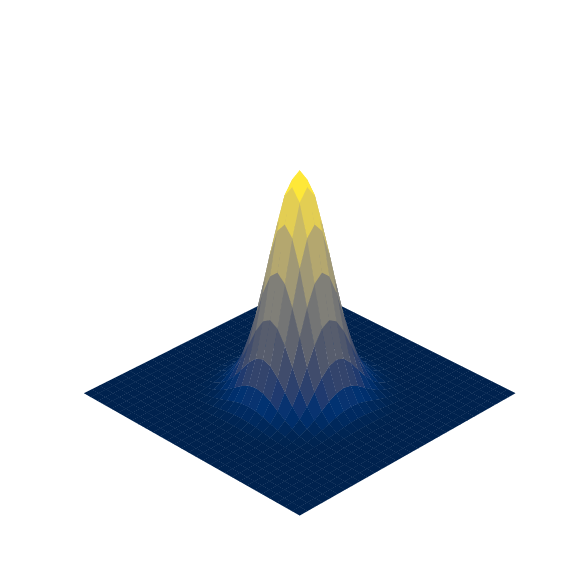}%
}
\hfil
\subfloat[]{\includegraphics[width=.85in]{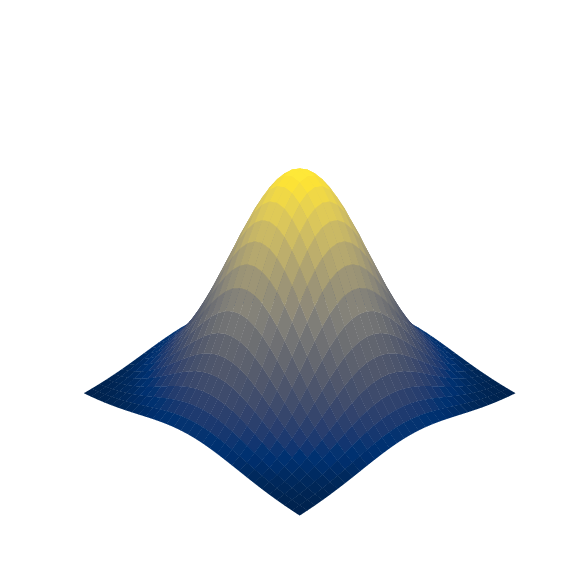}%
}
\hfil
\subfloat[]{\includegraphics[width=.85in]{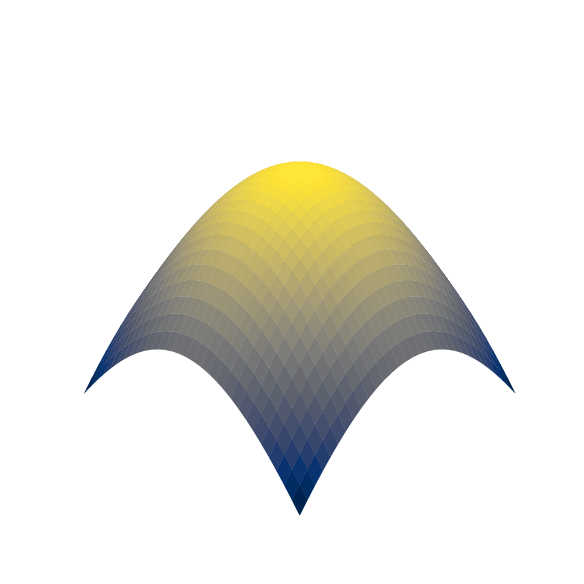}%
}
\caption{Examples of precomputed Gaussian filters. (a) $\sigma=2.98$. (b) $\sigma=6.85$. (c) $\sigma=20.0$.}
\label{fig:gaussians}
\end{figure}

Drawing from the suitability of the disk representation of trees, and the potential high precision of heatmap-based approaches, we propose a new approach for tree detection that is based only on a predicted heatmap. By generating Gaussian kernels with standard deviations depending on the crown diameter (similarly to the variable size Gaussians in CenterNet), we include the size information directly in the heatmap. During training, we optimize the model with a simple L1 loss to produce heatmaps. During inference, we identify peaks in the heatmap, and deduce size information by computing the normalized correlation between the local patch around each peak and a set of pre-computed isotropic Gaussian kernels, with evenly spaced standard deviations on a logarithmic scale (see Figure~\ref{fig:gaussians}):

\begin{equation}
    G = \Big\{\frac{1}{2\pi\sigma_i^2} e^{-\frac{x^2+y^2}{2\sigma_i^2}} \Big| \sigma_i = 10^p, p \in [p_{\text{min}}, p_{\text{max}}]\Big\}
\end{equation}

The Gaussian kernel with the highest correlation will give the predicted size of the corresponding tree. Compared to CenterNet, this has the advantage of removing the size prediction network, making the optimization process more straightforward. We will refer to this framework as \textit{Heatmap}.

\section{Experiments}
\label{sec:experiments}

\begin{figure}[t]
  \centering
   \includegraphics[width=\linewidth]{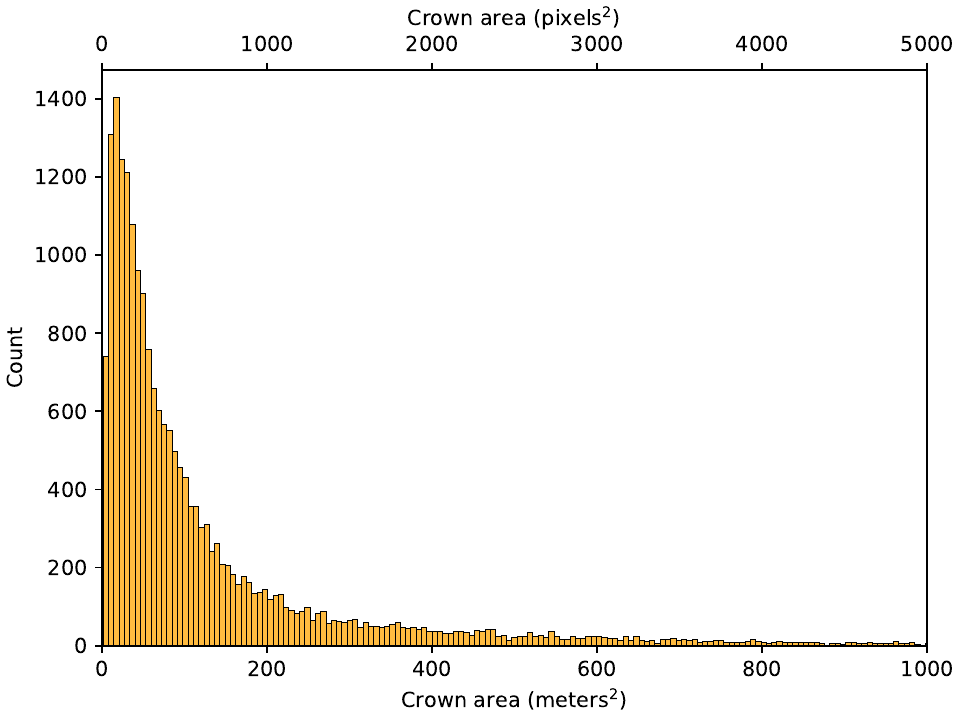}
   \caption{Distribution of crown areas in the Denmark dataset, in pixels and meters.}
   \label{fig:cdhist}
\end{figure}

We trained and evaluated the methods and architectures mentioned in Section~\ref{sec:models}, using two manually annotated datasets. First dataset contains 23.6k trees over different areas in Denmark, labeled on RGB+NIR aerial imagery \citep{li_deep_2023} at 20cm ground resolution. Second dataset contains 98.8k trees in Rwanda, labeled on RGB aerial imagery \citep{mugabowindekwe_nation-wide_2023} at 25cm ground resolution. The datasets include nordic forests, urban areas, savannas, farmlands, plantations, and rainforest. The labels were done for nation-wide mapping projects, accordingly they voluntarily cover difficult situations with dense areas or trees that can be easily mistaken for bushes or background elements. For an evaluation of the datasets with field data we refer to the corresponding papers. We train and validate on three different, randomly selected splits of approx. 60/20\% of the labeled data, and test on the remaining, fixed test split. We apply random vertical and horizontal flipping, random cropping and random resizing during training to artificially augment the dataset. Metrics (except for counting MAE) are computed with $\gamma = \{0.5, 1.0, 2.0\}$. In practice, that corresponds to thresholds of \{21cm, 42cm, 84cm\} / \{1px, 2px, 4px\} for the smallest tree in our dataset and \{12.8m, 25.6m, 51.2m\} / \{64px, 128, 256px\} for the biggest tree. Figure~\ref{fig:cdhist} plots the distribution of tree crown areas. 

Tree centers are derived as the geometric centers of labelled polygons. This might produce situations where the center lies outside of the polygon.
This situation occurs once in the Denmark dataset and 36 times in the Rwanda dataset, rarely enough to be neglected.

\begin{figure}[t]
  \centering
   \includegraphics[width=\linewidth]{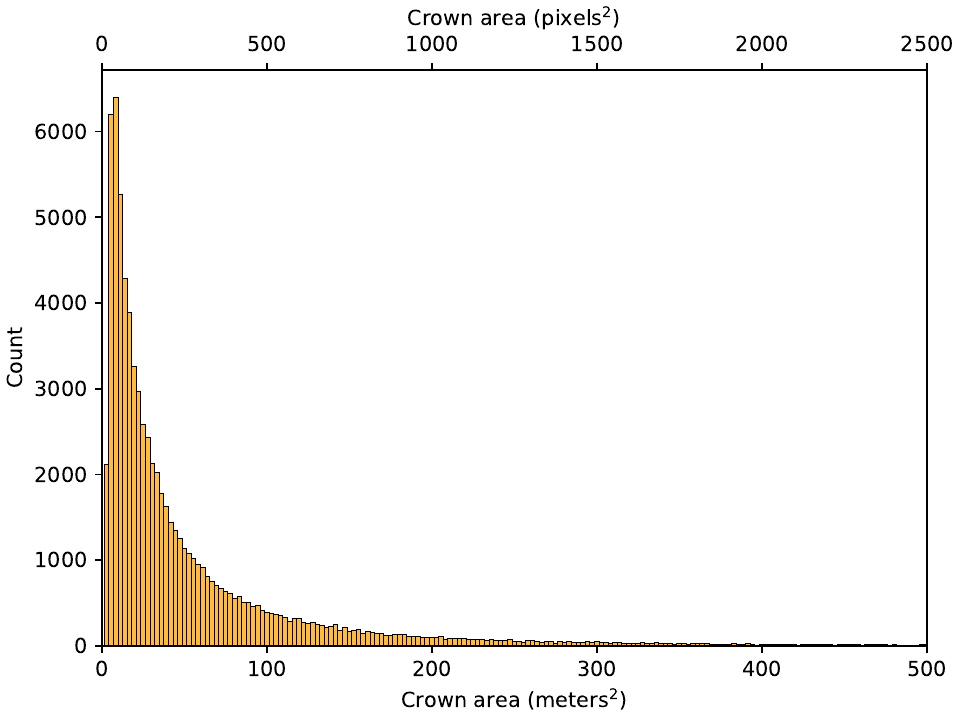}
   \caption{Distribution of crown areas in the Rwanda dataset, in pixels and meters.}
   \label{fig:cdhist_rwanda}
\end{figure}

Tables~\ref{tab:detresults}, \ref{tab:locresults} and \ref{tab:caresults} display the results for detecting, localizing and measuring trees respectively.

\subsection{Detection}

\begin{table*}
\caption{Tree detection benchmarking results. For each $\gamma$ (distance threhsold) we measure precision, recall and $\mathrm{F_1}$ score. Values are averaged accross 5 runs with different train/val splits, we indicate mean and standard deviation. Best value in bold.}
\label{tab:detresults}
\centering
  \begin{adjustbox}{width=\textwidth}
  \begin{tabular}{|cc|c|ccc|ccc|ccc|}
    \hline
    \multirow{2}{*}{Framework} & \multirow{2}{*}{Architecture}  & \multirow{2}{*}{\begin{tabular}{c} Counting nMAE \\ (\%) $\downarrow$ \end{tabular}}  & \multicolumn{3}{c|}{$\gamma = 0.5$} & \multicolumn{3}{|c|}{$\gamma = 1$} & \multicolumn{3}{|c|}{$\gamma = 2$}\\
     & & & Precision $\uparrow$ & Recall $\uparrow$ & $\mathrm{bF_1}$ $\uparrow$ & Precision $\uparrow$ & Recall $\uparrow$ & $\mathrm{bF_1}$ $\uparrow$ & Precision $\uparrow$ & Recall $\uparrow$ & $\mathrm{bF_1}$ $\uparrow$ \\
    \hline
    \multicolumn{12}{|c|}{\large Denmark} \\
    \hline
    \multirow{1}{*}{Segmentation} & UNet-R50 & $30.7\scriptstyle{\pm2.9}$ & $31.2\scriptstyle{\pm0.5}$ & $30.2\scriptstyle{\pm2.3}$ & $29.6\scriptstyle{\pm1.3}$ & $44.7\scriptstyle{\pm1.4}$ & $42.9\scriptstyle{\pm3.9}$ & $42.1\scriptstyle{\pm2.6}$ & $54.7\scriptstyle{\pm1.7}$ & $57.2\scriptstyle{\pm3.8}$ & $52.9\scriptstyle{\pm2.2}$
\\
    \hline
    \multirow{1}{*}{CenterNet} & UNet-R50 & $55.2\scriptstyle{\pm2.8}$ & $29.2\scriptstyle{\pm2.9}$ & $31.6\scriptstyle{\pm4.3}$ & $28.5\scriptstyle{\pm2.7}$ & $43.1\scriptstyle{\pm4.8}$ & $44.9\scriptstyle{\pm6.2}$ & $40.7\scriptstyle{\pm3.6}$ & $51.4\scriptstyle{\pm5.6}$ & $55.0\scriptstyle{\pm6.6}$ & $48.3\scriptstyle{\pm3.8}$
\\
    \hline
    \multirow{1}{*}{Heatmap} & UNet-R50 & $33.4\scriptstyle{\pm3.2}$ & $33.8\scriptstyle{\pm1.6}$ & $29.9\scriptstyle{\pm3.7}$ & $30.1\scriptstyle{\pm3.0}$ & $48.1\scriptstyle{\pm0.4}$ & $41.6\scriptstyle{\pm5.1}$ & $41.8\scriptstyle{\pm3.5}$ & $54.9\scriptstyle{\pm1.2}$ & $49.3\scriptstyle{\pm6.3}$ & $48.1\scriptstyle{\pm4.1}$
\\
     \hline
    \multirow{1}{*}{Point proposal} & P2P & $26.4\scriptstyle{\pm1.6}$ & $34.2\scriptstyle{\pm2.5}$ & $29.0\scriptstyle{\pm1.9}$ & $30.4\scriptstyle{\pm1.6}$ & $56.0\scriptstyle{\pm3.7}$ & $47.9\scriptstyle{\pm2.5}$ & $\mathbf{49.6}\scriptstyle{\pm1.8}$ & $67.7\scriptstyle{\pm3.7}$ & $61.0\scriptstyle{\pm2.0}$ & $\mathbf{60.6}\scriptstyle{\pm1.6}$
\\
    \hline
    \multirow{1}{*}{Box proposal} & FasterRCNN & $39.9\scriptstyle{\pm4.6}$ & $34.9\scriptstyle{\pm1.5}$ & $34.7\scriptstyle{\pm0.9}$ & $\mathbf{32.7}\scriptstyle{\pm1.1}$ & $50.0\scriptstyle{\pm2.9}$ & $52.4\scriptstyle{\pm2.3}$ & $47.6\scriptstyle{\pm2.0}$ & $58.0\scriptstyle{\pm3.1}$ & $64.6\scriptstyle{\pm2.2}$ & $56.0\scriptstyle{\pm1.6}$ 

 \\
    \hline
    \multicolumn{12}{|c|}{\large Rwanda} \\
    \hline
    \multirow{1}{*}{Segmentation} & UNet-R50 & $64.4\scriptstyle{\pm1.0}$ & $14.3\scriptstyle{\pm0.6}$ & $16.3\scriptstyle{\pm1.0}$ & $13.7\scriptstyle{\pm0.8}$ & $34.6\scriptstyle{\pm1.7}$ & $41.8\scriptstyle{\pm3.1}$ & $34.3\scriptstyle{\pm2.2}$ & $51.6\scriptstyle{\pm1.6}$ & $65.9\scriptstyle{\pm2.8}$ & $52.0\scriptstyle{\pm2.1}$ \\
    \hline
    \multirow{1}{*}{CenterNet} & UNet-R50 & $75.9\scriptstyle{\pm4.0}$ & $13.5\scriptstyle{\pm0.4}$ & $14.2\scriptstyle{\pm1.3}$ & $12.1\scriptstyle{\pm0.6}$ & $34.5\scriptstyle{\pm0.3}$ & $38.9\scriptstyle{\pm4.4}$ & $31.9\scriptstyle{\pm2.3}$ & $51.7\scriptstyle{\pm0.9}$ & $60.2\scriptstyle{\pm6.2}$ & $48.0\scriptstyle{\pm2.8}$ \\
    \hline
    \multirow{1}{*}{Heatmap} & UNet-R50 & $78.5\scriptstyle{\pm7.6}$ & $12.6\scriptstyle{\pm1.2}$ & $13.7\scriptstyle{\pm1.3}$ & $11.7\scriptstyle{\pm1.2}$ & $32.8\scriptstyle{\pm3.0}$ & $38.6\scriptstyle{\pm4.8}$ & $31.4\scriptstyle{\pm4.0}$ & $50.8\scriptstyle{\pm3.4}$ & $63.8\scriptstyle{\pm5.5}$ & $49.6\scriptstyle{\pm5.1}$ \\
     \hline
    \multirow{1}{*}{Point proposal} & P2P & $68.7\scriptstyle{\pm7.7}$ & $13.9\scriptstyle{\pm1.1}$ & $16.9\scriptstyle{\pm1.9}$ & $\mathbf{13.8}\scriptstyle{\pm1.2}$ & $34.8\scriptstyle{\pm1.8}$ & $44.9\scriptstyle{\pm3.4}$ & $\mathbf{35.3}\scriptstyle{\pm1.9}$ & $52.3\scriptstyle{\pm2.7}$ & $70.7\scriptstyle{\pm2.3}$ & $\mathbf{53.6}\scriptstyle{\pm1.9}$ \\
    \hline
    \multirow{1}{*}{Box proposal} & FasterRCNN & $78.1\scriptstyle{\pm5.9}$ & $14.1\scriptstyle{\pm0.5}$ & $17.5\scriptstyle{\pm0.8}$ & $13.6\scriptstyle{\pm0.5}$ & $33.3\scriptstyle{\pm1.2}$ & $45.1\scriptstyle{\pm1.3}$ & $33.6\scriptstyle{\pm1.1}$ & $49.0\scriptstyle{\pm2.1}$ & $68.2\scriptstyle{\pm1.5}$ & $49.5\scriptstyle{\pm1.7}$ \\
    \hline
    
\end{tabular}
\end{adjustbox}
\end{table*}

The detection Dice score naturally improves with higher values of $\gamma$, \ie with a more tolerant positive definition. 

Surprisingly, the segmentation approach, which has access to the most detailed information (pixel labels), does not perform significantly better than other methods. The P2P approach achieves the best bF1 score on 5 out of 6 setups, with a large margin. We hypothesize that having the 1-to-1 matching process built-in for optimization helps getting better detections. 

Heatmap-based methods perform comparatively better at low $\gamma$ values. At $\gamma$ = 0.5, no method produces a significantly higher score, and bF1 scores are considerably lower. This highlights the difficulty of precise individual tree mapping.

\subsection{Localization} 

\begin{table}
\caption{Localization benchmarking results. For each $\gamma$ (distance threhsold) we measure localization (Loc.) mean absolute error.}
\label{tab:locresults}
\centering
  \begin{adjustbox}{width=\linewidth}
  \begin{tabular}{|cc|c|c|c|}
    \hline
    \multirow{2}{*}{Framework} & \multirow{2}{*}{Architecture} & \multicolumn{1}{c|}{$\gamma = 0.5$} & \multicolumn{1}{|c|}{$\gamma = 1$} & \multicolumn{1}{|c|}{$\gamma = 2$}\\
     & & Loc. RMSE (m) $\downarrow$ & Loc. RMSE (m) $\downarrow$ & Loc. RMSE (m) $\downarrow$ \\
    \hline
    \multicolumn{5}{|c|}{\large Denmark} \\
    \hline
    \multirow{1}{*}{Segmentation} & UNet-R50 & $0.48\scriptstyle{\pm0.02}$ & $0.68\scriptstyle{\pm0.03}$ & $1.24\scriptstyle{\pm0.1}$\\
    \hline
    \multirow{1}{*}{\begin{tabular}{c} CenterNet  \end{tabular}} & UNet-R50 & $0.47\scriptstyle{\pm0.02}$ & $0.7\scriptstyle{\pm0.07}$ & $1.18\scriptstyle{\pm0.12}$ \\
     \hline
    \multirow{1}{*}{Heatmap} & UNet-R50 & $\mathbf{0.45}\scriptstyle{\pm0.0}$ & $\mathbf{0.66}\scriptstyle{\pm0.02}$ & $\mathbf{1.14}\scriptstyle{\pm0.02}$ \\
     \hline
    \multirow{1}{*}{Point proposal} & P2P & $0.53\scriptstyle{\pm0.03}$ & $0.78\scriptstyle{\pm0.02}$ & $1.26\scriptstyle{\pm0.03}$\\
    \hline
    \multirow{1}{*}{Box proposal} & FasterRCNN & $0.49\scriptstyle{\pm0.01}$ & $0.69\scriptstyle{\pm0.01}$ & $1.17\scriptstyle{\pm0.03}$ \\
    \hline
    \multicolumn{5}{|c|}{\large Rwanda} \\
    \hline
    \multirow{1}{*}{Segmentation} & UNet-R50 & $0.84\scriptstyle{\pm0.04}$ & $1.27\scriptstyle{\pm0.05}$ & $1.77\scriptstyle{\pm0.05}$ \\
    \hline
    \multirow{1}{*}{\begin{tabular}{c} CenterNet  \end{tabular}} & UNet-R50 & $0.8\scriptstyle{\pm0.05}$ & $1.23\scriptstyle{\pm0.05}$ & $1.75\scriptstyle{\pm0.06}$ \\
     \hline
    \multirow{1}{*}{Heatmap} & UNet-R50 & $\mathbf{0.68}\scriptstyle{\pm0.1}$ & $\mathbf{1.06}\scriptstyle{\pm0.16}$ & $\mathbf{1.53}\scriptstyle{\pm0.18}$ \\
     \hline
    \multirow{1}{*}{Point proposal} & P2P & $0.85\scriptstyle{\pm0.04}$ & $1.29\scriptstyle{\pm0.02}$ & $1.86\scriptstyle{\pm0.11}$ \\
    \hline
    \multirow{1}{*}{Box proposal} & FasterRCNN & $0.8\scriptstyle{\pm0.03}$ & $1.19\scriptstyle{\pm0.05}$ & $1.71\scriptstyle{\pm0.03}$ \\
    \hline
\end{tabular}
\end{adjustbox}
\end{table}

Our proposed Heatmap framework achieves the lowest localization error at all $\gamma$ values. Compared to CenterNet, this shows the advantage of fusing the outputs into a single heatmap, simplifying the optimization space and leading to more accurate models.

\subsection{Crown area estimation} 

\begin{table}
\caption{Crown area estimation benchmarking results. For each $\gamma$ (distance threhsold) we measure crown diameter (CA) mean absolute error.}
\label{tab:caresults}
\centering
  \begin{adjustbox}{width=\linewidth}
  \begin{tabular}{|cc|c|c|c|}
    \hline
    \multirow{2}{*}{Framework} & \multirow{2}{*}{Architecture} & \multicolumn{1}{c|}{$\gamma = 0.5$} & \multicolumn{1}{|c|}{$\gamma = 1$} & \multicolumn{1}{|c|}{$\gamma = 2$}\\
     & & CA RMSE ($m^2$) $\downarrow$ & CA RMSE ($m^2$) $\downarrow$ & CA RMSE ($m^2$) $\downarrow$ \\
    \hline
    \multicolumn{5}{|c|}{\large Denmark} \\
    \hline
    \multirow{1}{*}{Segmentation} & UNet-R50 & $16.35\scriptstyle{\pm0.81}$ & $17.38\scriptstyle{\pm0.68}$ & $20.87\scriptstyle{\pm0.63}$ \\
    \hline
    \multirow{1}{*}{\begin{tabular}{c} CenterNet  \end{tabular}} & UNet-R50 & $14.57\scriptstyle{\pm1.19}$ & $16.0\scriptstyle{\pm0.43}$ & $\mathbf{19.73}\scriptstyle{\pm1.06}$  \\
     \hline
    \multirow{1}{*}{Heatmap} & UNet-R50 & $19.29\scriptstyle{\pm1.11}$ & $19.68\scriptstyle{\pm0.76}$ &
$21.68\scriptstyle{\pm0.42}$ \\
     \hline
    \multirow{1}{*}{Point proposal} & P2P & $\mathbf{13.89}\scriptstyle{\pm0.57}$ & $\mathbf{14.45}\scriptstyle{\pm0.36}$ & $20.48\scriptstyle{\pm1.03}$\\
    \hline
    \multirow{1}{*}{Box proposal} & FasterRCNN & $14.69\scriptstyle{\pm0.27}$ & $15.95\scriptstyle{\pm0.35}$ & $20.55\scriptstyle{\pm0.43}$\\
    \hline
    \multicolumn{5}{|c|}{\large Rwanda} \\
    \hline
    \multirow{1}{*}{Segmentation} & UNet-R50 & $36.4\scriptstyle{\pm1.2}$ & $31.61\scriptstyle{\pm0.68}$ & $34.58\scriptstyle{\pm1.05}$ \\
    \hline
    \multirow{1}{*}{\begin{tabular}{c} CenterNet  \end{tabular}} & UNet-R50 & $39.45\scriptstyle{\pm0.63}$ & $34.31\scriptstyle{\pm0.9}$ & $36.77\scriptstyle{\pm1.14}$ \\
     \hline
    \multirow{1}{*}{Heatmap} & UNet-R50 & $37.64\scriptstyle{\pm8.21}$ & $33.55\scriptstyle{\pm6.36}$ & $35.59\scriptstyle{\pm6.41}$ \\
     \hline
    \multirow{1}{*}{Point proposal} & P2P & $34.17\scriptstyle{\pm2.2}$ & $34.48\scriptstyle{\pm5.31}$ & $38.87\scriptstyle{\pm5.81}$ \\
    \hline
    \multirow{1}{*}{Box proposal} & FasterRCNN & $\mathbf{29.1}\scriptstyle{\pm1.06}$ & $\mathbf{28.82}\scriptstyle{\pm1.19}$ & $\mathbf{33.33}\scriptstyle{\pm1.07}$ \\
    \hline
\end{tabular}
\end{adjustbox}
\end{table}

The Box proposal framework with FasterRCNN achieves the lowest crown area estimation error on the Rwanda dataset, and competitive errors on the Denmark dataset. 

Surprisingly, the Segmentation approach does not perform better, despite having access to more precise shape information. The circular and rectangular approximations done by other methods to reduce the problem to a single value prediction behave on par or better, showing the benefit of simplifying the problem to learn better models.

\subsection{Feature extractors}

\begin{table}
\caption{Comparison of feature extractors for the segmentation and heatmap frameworks, on the Denmark dataset.}
\label{tab:featureextractor}
\centering
  \begin{adjustbox}{width=\linewidth}
  \begin{tabular}{|cc|c|ccc|}
    \hline
    \multirow{2}{*}{Framework} & \multirow{2}{*}{Architecture}  & \multirow{2}{*}{\begin{tabular}{c} Counting MAE \\ (\%) $\downarrow$ \end{tabular}}  & \multicolumn{3}{c|}{$\mathrm{bF_1}$} \\
     & & & $\gamma = 0.5$ & $\gamma = 1$ & $\gamma = 2$\\
    \hline
    \hline
\multirow{4}{*}{Segmentation} & UNet-R50 & $30.7\scriptstyle{\pm2.9}$ & $29.6\scriptstyle{\pm1.3}$ & $42.1\scriptstyle{\pm2.6}$ & $52.9\scriptstyle{\pm2.2}$
\\
    & DeepLabV3 & $33.0\scriptstyle{\pm6.2}$ & $26.8\scriptstyle{\pm1.2}$ & $40.4\scriptstyle{\pm0.8}$ & $51.1\scriptstyle{\pm1.3}$
\\
     & SegFormer & $32.8\scriptstyle{\pm2.3}$ & $24.8\scriptstyle{\pm2.4}$ & $40.3\scriptstyle{\pm3.0}$ & $52.4\scriptstyle{\pm3.1}$
\\
     & TransUNet & $31.8\scriptstyle{\pm3.6}$ & $29.2\scriptstyle{\pm1.1}$ & $41.8\scriptstyle{\pm2.3}$ & $51.9\scriptstyle{\pm2.4}$
\\
    \hline
    \multirow{4}{*}{Heatmap} & UNet-R50 & $33.4\scriptstyle{\pm3.2}$ & $30.1\scriptstyle{\pm3.0}$ & $41.8\scriptstyle{\pm3.5}$ & $48.1\scriptstyle{\pm4.1}$
\\
         & DeepLabV3 & $37.2\scriptstyle{\pm5.5}$ & $30.9\scriptstyle{\pm1.5}$ & $42.8\scriptstyle{\pm2.1}$ & $51.2\scriptstyle{\pm1.1}$
\\
         & SegFormer & $36.4\scriptstyle{\pm4.6}$ & $27.9\scriptstyle{\pm2.9}$ & $42.0\scriptstyle{\pm2.4}$ & $50.4\scriptstyle{\pm1.5}$
\\
     & TransUNet & $36.0\scriptstyle{\pm1.4}$ & $29.9\scriptstyle{\pm1.2}$ & $40.0\scriptstyle{\pm2.4}$ & $46.1\scriptstyle{\pm3.1}$
\\
    \hline
\end{tabular}
\end{adjustbox}
\end{table}

Heatmap-based approaches can be implemented with many different backbones without the need for extensive redesign. Segmentation for example is a well-studied task in computer vision and many architectures have been proposed to address this. Those various architectures can also be repurposed for our proposed Heatmap approach. Criteria of choice are mainly the number of parameters, architectural choices, and output resolution. We implemented four popular feature extractors (see \ref{appx:fe} for more details). with the Segmentation and Heatmap methods, and compared them on our benchmark.

The results (Table~\ref{tab:featureextractor}) confirm that UNet remains a strong baseline. The other architectures, which were all originally designed for pixel classification, do not perform better here, despite having more parameters and using advanced neural architectures.

\subsection{Area-based evaluation}

Our proposed object-centric evaluation framework is voluntarily oriented towards metrics that relate to individual scalar values, here tree position and crown diameter. We derive those values from manually labelled crown polygons, thereby discarding the information of crown shape. However, evaluating how well methods predict the canopy cover has applicative interests such as estimating canopy cover over areas or creating visually consistent maps.

Segmentation methods, the gold standard for predicting shape, are usually trained and evaluated with area based metrics, such as the IoU (Intersection over Union). They use patch-level IoU as a proxy for shape prediction at individual level, relying on a carefully tuned instance separation step. The other methods we describe in Section~\ref{subsec:frameworks} predict size as a single value (CA), which involves approximations that inherently limit how well they can map canopy cover.

After matching trees with predictions, we can study how well methods map individual tree cover by comparing the predicted shape (an individual segment for the Segmentation framework, a disk for other frameworks) with the labelled polygon. The Heatmap framework has the lowest upper bound, due to the discretization of Gaussian sizes. The Box proposal framework, estimating the radius from bounding box height and width, also has a slightly different bound compared to the CenterNet and Point proposal frameworks, which directly output radius.

Unsurprisingly, the Segmentation approach has a good patch IoU, the metric it is optimized for. But individual IoU reveals a different behavior, with the Segmentation approach being in fact the worst. This indicates that patch-level IoU does not necessarily correlate with individual IoU, and highlights the impact of the arduous instance separation step in post-processing. The observed differences between patch-level and individual IoU performance are a result of the one-to-one mapping enforced in our evaluation. The detection framworks have the best individual IoU, which is consistent with their performance for CA estimation (Table~\ref{tab:caresults}). We note that the upper bounds induced by the circular approximation do not seem to correlate with measured IoUs, suggesting that other factors such as the amount of training data or the optimization process are more limiting.

Individual IoU in Table~\ref{tab:areabased} correlates well with the individual localization and CA measures (Tables~\ref{tab:locresults} \& \ref{tab:caresults}), but does not guarantee accurate counting or detection. Similarly, patch IoU favors Segmentation but does not say much about other aspects of performance, concealing local effects by a single compound metric, which may, on the other hand, be more relevant when results are aggregated over a larger area.

\begin{table}[]
    \caption{Area-based metrics: upper bounds induced by design (circular approximation) and measured performance after training models. We indicate patch-level IoU for reference. Individual IoU is measured at $\gamma = 0.5$, with one-to-one matching. }
    \label{tab:areabased}
    \centering
    \begin{adjustbox}{width=\linewidth}
    \begin{tabular}{|c|c|c|c|c|c|c|c|c|}
        \hline
         & \multicolumn{2}{|c|}{Individual} & Patch \\
        Method & IoU upper bound (\%) & IoU measured (\%) & IoU measured (\%)\\
        \hline
        \multicolumn{6}{|c|}{Denmark} \\
        \hline
        Segmentation & $100.0\scriptstyle{\pm0.0}$ & $49.9\scriptstyle{\pm4.9}$ & $36.2\scriptstyle{\pm2.4}$\\
        CenterNet & $75.9\scriptstyle{\pm9.0}$ & $56.7\scriptstyle{\pm4.5}$ & $30.6\scriptstyle{\pm4.6}$\\
        Heatmap & $71.9\scriptstyle{\pm11.7}$ & $58.1\scriptstyle{\pm1.1}$ & $30.2\scriptstyle{\pm5.2}$\\
        Point proposal & $75.9\scriptstyle{\pm9.0}$ & $58.8\scriptstyle{\pm2.6}$ & $36.9\scriptstyle{\pm1.1}$\\
        Box proposal & $74.7\scriptstyle{\pm9.2}$ & $56.2\scriptstyle{\pm2.4}$ & $32.6\scriptstyle{\pm1.5}$\\
        \hline
        \multicolumn{6}{|c|}{Rwanda} \\
        \hline
        Segmentation & $100.0\scriptstyle{\pm0.0}$ & $43.4\scriptstyle{\pm1.8}$ & $29.8\scriptstyle{\pm5.7}$\\
        CenterNet & $70.9\scriptstyle{\pm13.2}$ & $46.6\scriptstyle{\pm1.3}$ & $27.2\scriptstyle{\pm4.6}$\\
        Heatmap & $68.2\scriptstyle{\pm14.2}$ & $45.1\scriptstyle{\pm2.5}$ & $28.7\scriptstyle{\pm4.7}$\\
        Point proposal & $70.9\scriptstyle{\pm13.2}$ & $49.3\scriptstyle{\pm2.4}$ & $26.4\scriptstyle{\pm4.4}$\\
        Box proposal & $70.0\scriptstyle{\pm12.2}$ & $51.7\scriptstyle{\pm2.0}$ & $25.4\scriptstyle{\pm1.7}$\\
        \hline
    \end{tabular}
    \end{adjustbox}
\end{table}

\subsection{Ensembling}

A common technique to gain accuracy is to use various models or transformations on the input data, and merge the outputs. This is particularly interesting in the field of tree mapping, where annotation costs are high, and datasets are therefore relatively small. Setting aside a portion for validation or testing can result in the removal of valuable training data, potentially reducing final performance.. A simple solution is to train N models in a N-fold cross-validation setup, and use them together in an ensemble model for predictions. This way, the ensemble model sees all training and validation labels, and the merging operation can mitigate bad predictions due to unstable singe models.

Heatmap-based methods are excellent candidates for ensembling, the output heatmaps from different models or data augmentations can simply be merged pixel-wise, provided that the image transformations are known and can be reversed on the output. Anchor-based methods, in contrast, are tricky to ensemble, since they output lists of predictions attached to movable anchors. Concatenating the detections, even with subsequent NMS, inevitably leads to over-prediction. 

For heatmap-based methods, we implemented model ensembling by merging the outputs of the models trained on three different train/val splits, and data ensembling (also known as test-time augmentation, TTA) by merging the outputs obtained with flipping, rotation and rescaling transforms on the input images. We use a simple averaging operation on the heatmaps (after applying inverse transforms), before decoding depending on the framework (segmentation, CenterNet, proposed Heatmap detection approach). 

\begin{table}[]
    \caption{Tree detection results with model and data ensembling}
    \label{tab:ensemble}
    \centering
  \begin{adjustbox}{width=\linewidth}
  \begin{tabular}{|cc|c|ccc|}
    \hline
    \multirow{2}{*}{Framework} & \multirow{2}{*}{Architecture}  & \multirow{2}{*}{\begin{tabular}{c} Counting nMAE \\ (\%) $\downarrow$ \end{tabular}}  & \multicolumn{3}{c|}{$\mathrm{bF_1}$} \\
     & & & $\gamma = 0.5$ & $\gamma = 1$ & $\gamma = 2$\\
    \hline
    \multicolumn{6}{|c|}{Denmark} \\
    \hline
    Segmentation & \multirow{2}{*}{UNet-R50} & $27.9$ & $32.6$ & $44.1$ & $53.2$\\
     \textit{+ TTA} & & $27.6$ & $31.4$ & $43.7$ & $52.6$\\
    \hline
    CenterNet & \multirow{2}{*}{UNet-R50} & $38.0$ & $32.3$ & $43.6$ & $50.3$\\
    \textit{+ TTA} & & $33.6$ & $32.1$ & $41.8$ & $47.5$\\
    \hline
    Heatmap & \multirow{2}{*}{UNet-R50} & $46.2$ & $37.1$ & $\mathbf{49.3}$ & $\mathbf{55.9}$\\
    \textit{+ TTA} & & $36.1$ & $\mathbf{37.5}$ & $48.0$ & $53.4$\\
    \hline
    \multicolumn{6}{|c|}{Rwanda} \\
    \hline
    Segmentation & \multirow{2}{*}{UNet-R50} & $60.3$ & $13.0$ & $33.5$ & $51.9$\\
     \textit{+ TTA} & & $57.7$ & $13.8$ & $35.4$ & $52.4$\\
    \hline
    CenterNet & \multirow{2}{*}{UNet-R50} & $72.5$ & $13.0$ & $33.5$ & $51.2$\\
    \textit{+ TTA} & & $61.5$ & $13.2$ & $34.8$ & $53.6$\\
    \hline
    Heatmap & \multirow{2}{*}{UNet-R50} & $71.7$ & $17.3$ & $40.3$ & $54.9$\\
    \textit{+ TTA} & & $61.1$ & $\mathbf{18.9}$ & $\mathbf{43.7}$ & $\mathbf{57.0}$\\
    \hline
\end{tabular}
\end{adjustbox}
\end{table}

We observe a consistent but inequal boost in performance with ensembling, comparing to the average performance of individual models. CenterNet and Segmentation approaches produce +2\% and +7\% bF1 scores, respectively with model ensembling, and +3\% and +7\% with TTA. The Heatmap detection approach benefits significantly more from ensembling, with +24\% on average with model ensembling and +28\% with TTA. We argue that having a single output with a single loss function leads to better predicted Gaussians, making the averaged results smoother and thus benefitting more from ensembling.

\subsection{Qualitative results}

\begin{figure*}[!t]

\captionsetup[subfloat]{captionskip=-1.2em,justification=raggedleft,singlelinecheck=false,font=normalsize,labelfont={color=white,bf}}
\centering
\subfloat[]{\includegraphics[width=.24\linewidth]{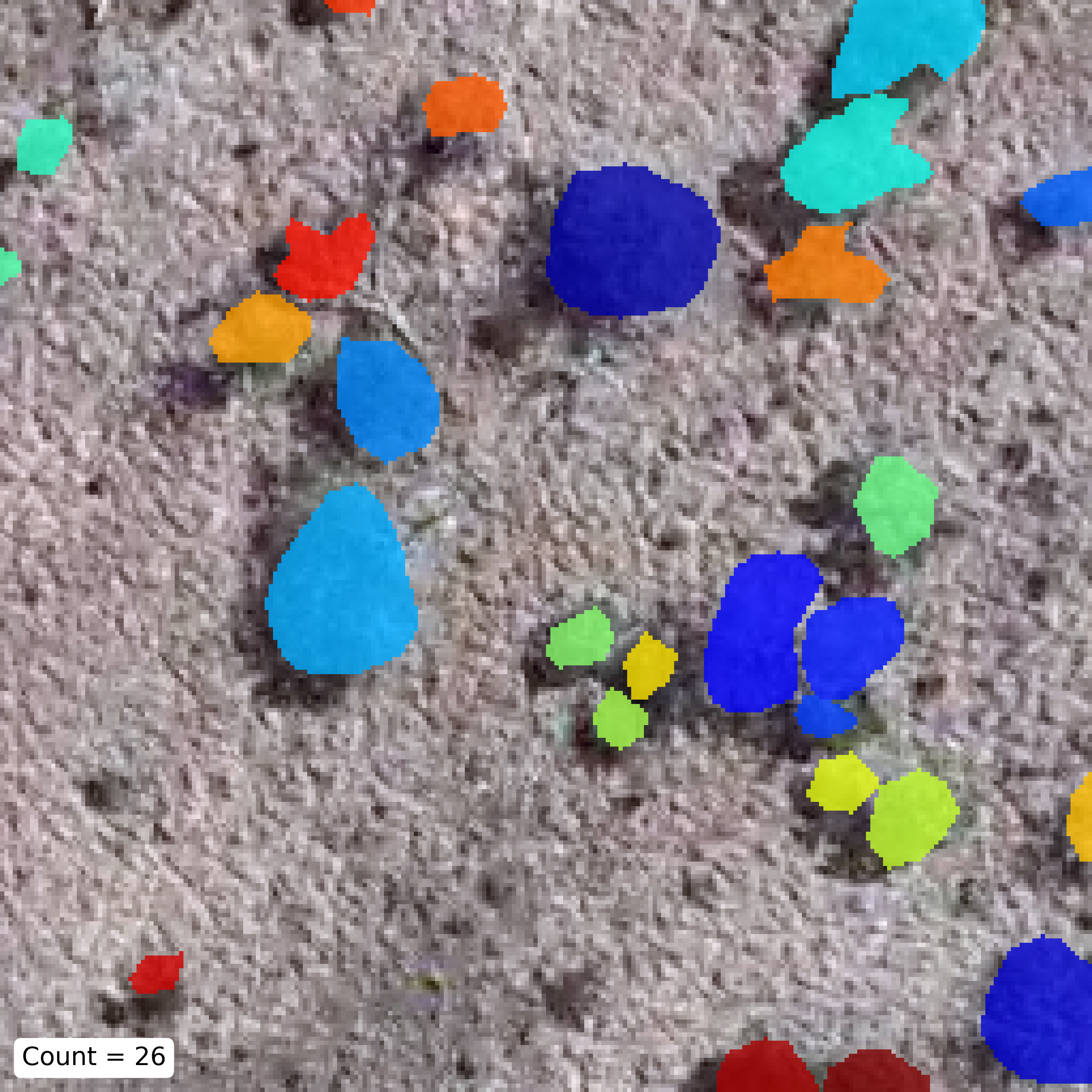}%
}
\hspace{.3ex}
\subfloat[]{\includegraphics[width=.24\linewidth]{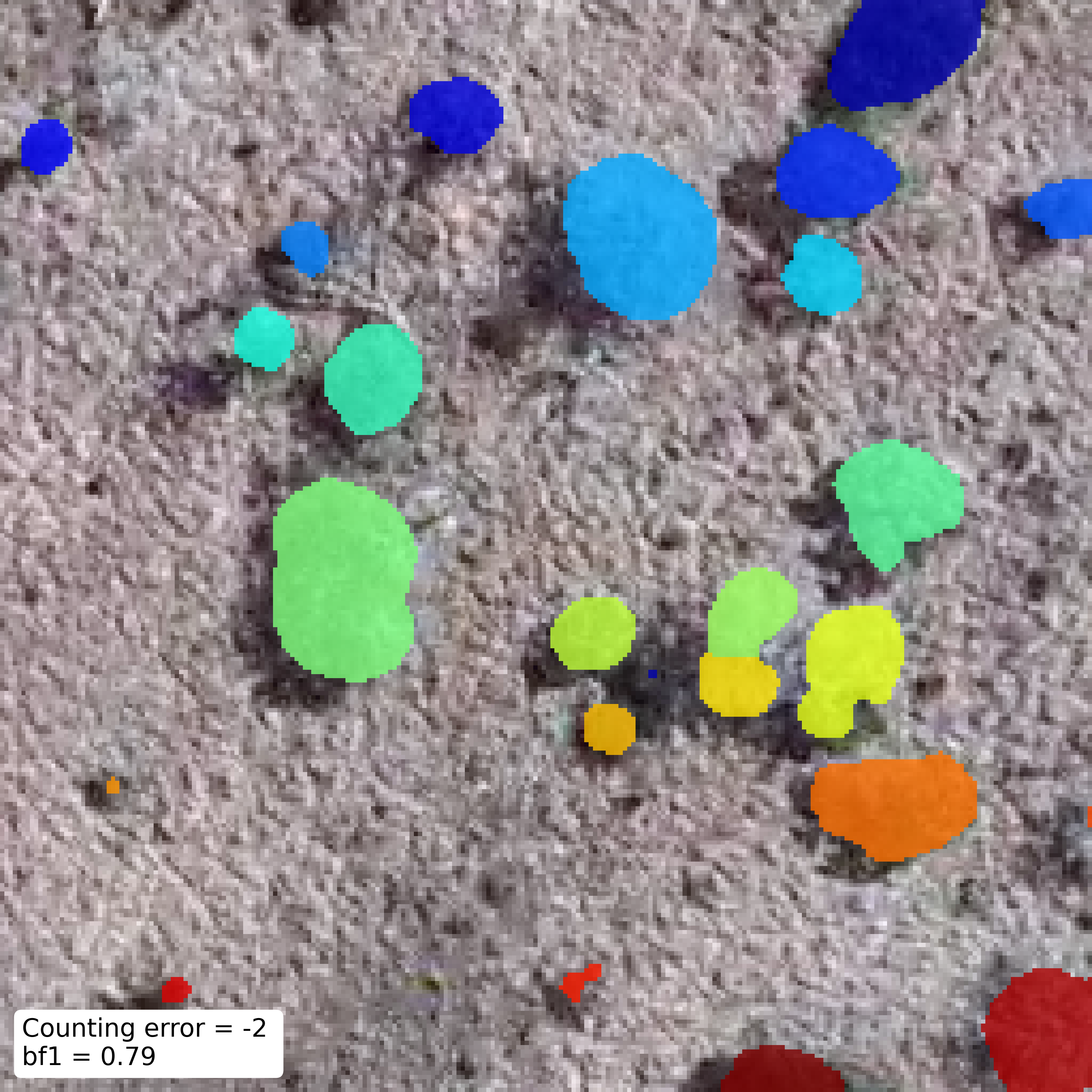}%
}
\hspace{.3ex}
\subfloat[]{\includegraphics[width=.24\linewidth]{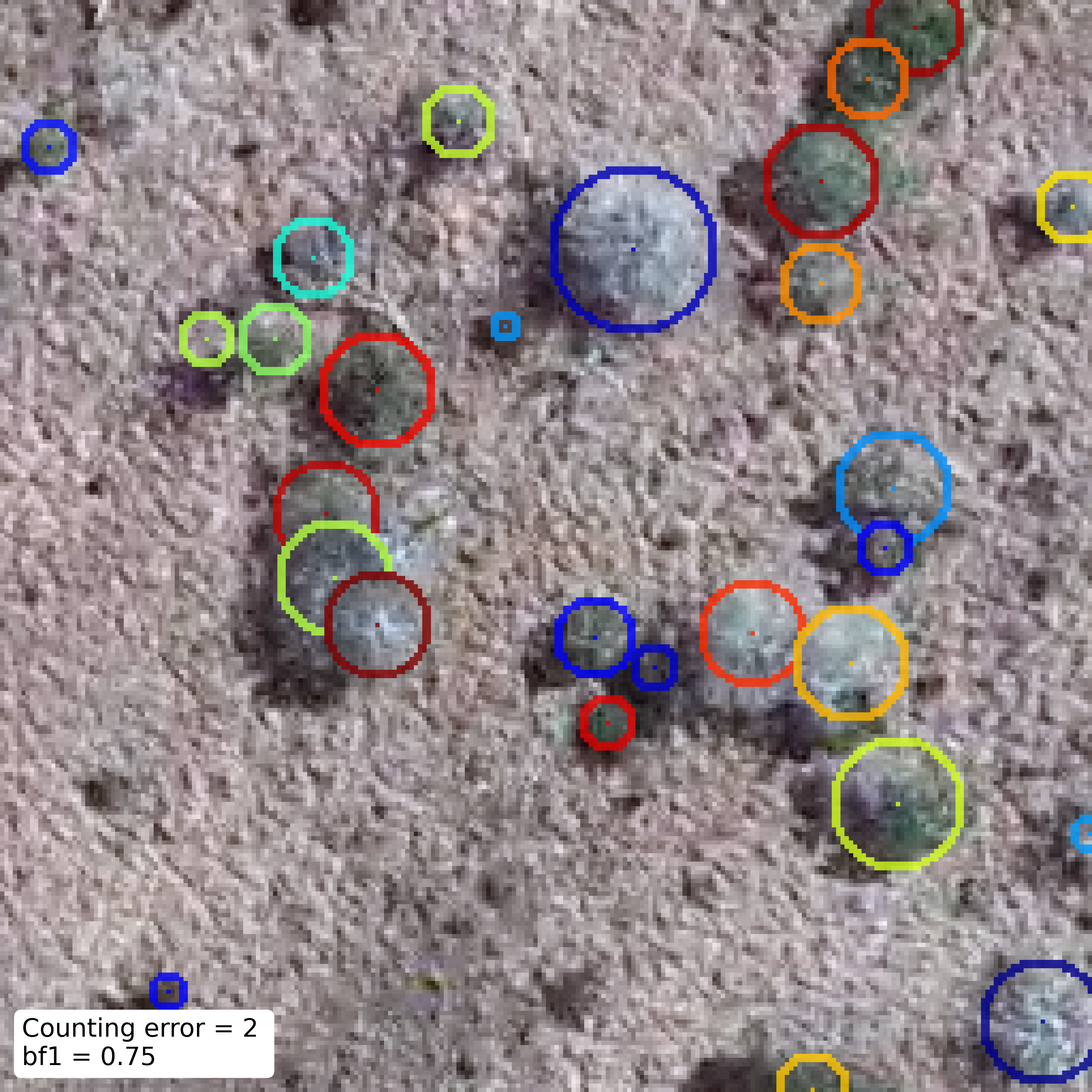}%
}
\\[1ex]
\subfloat[]{\includegraphics[width=.24\linewidth]{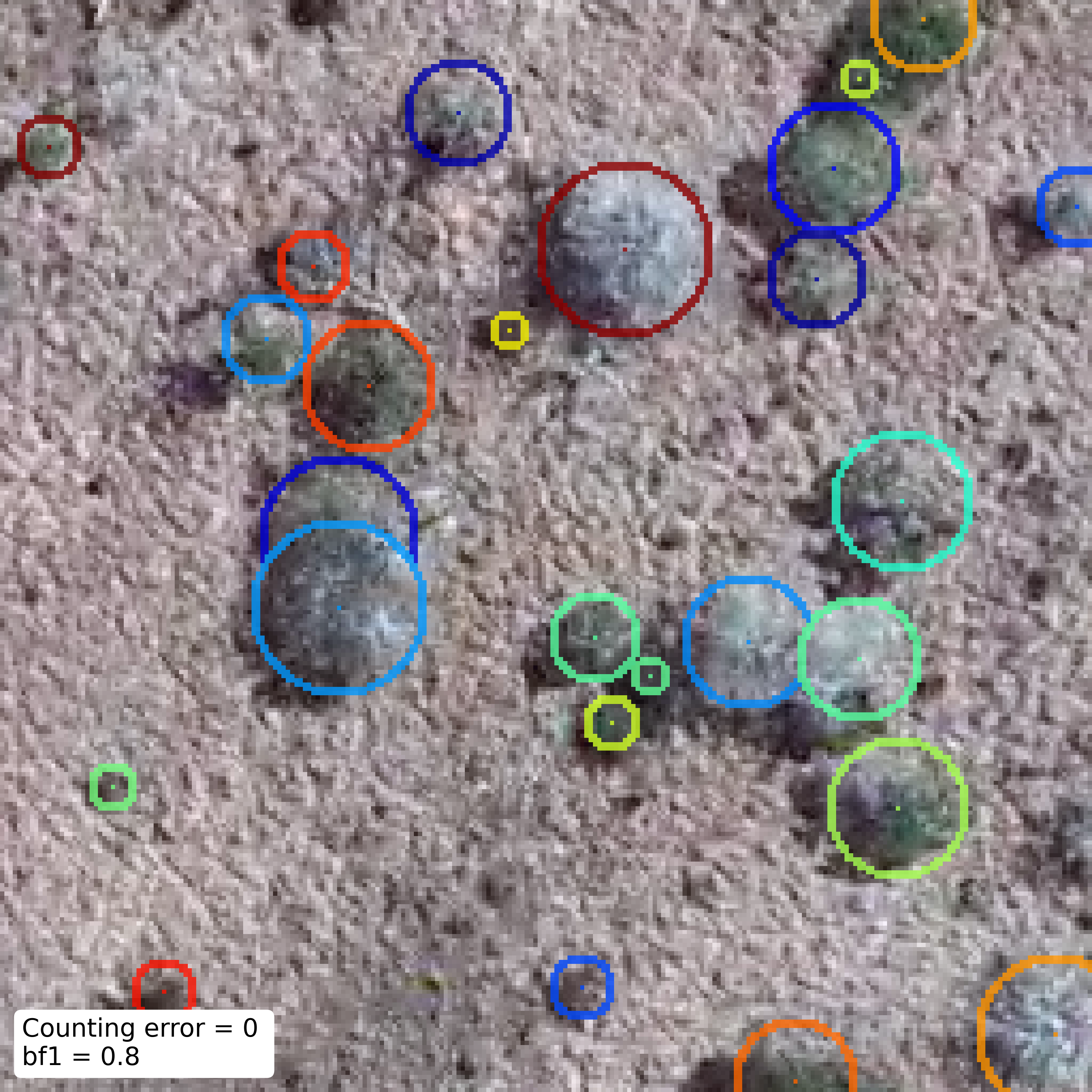}%
}
\hspace{.3ex}
\subfloat[]{\includegraphics[width=.24\linewidth]{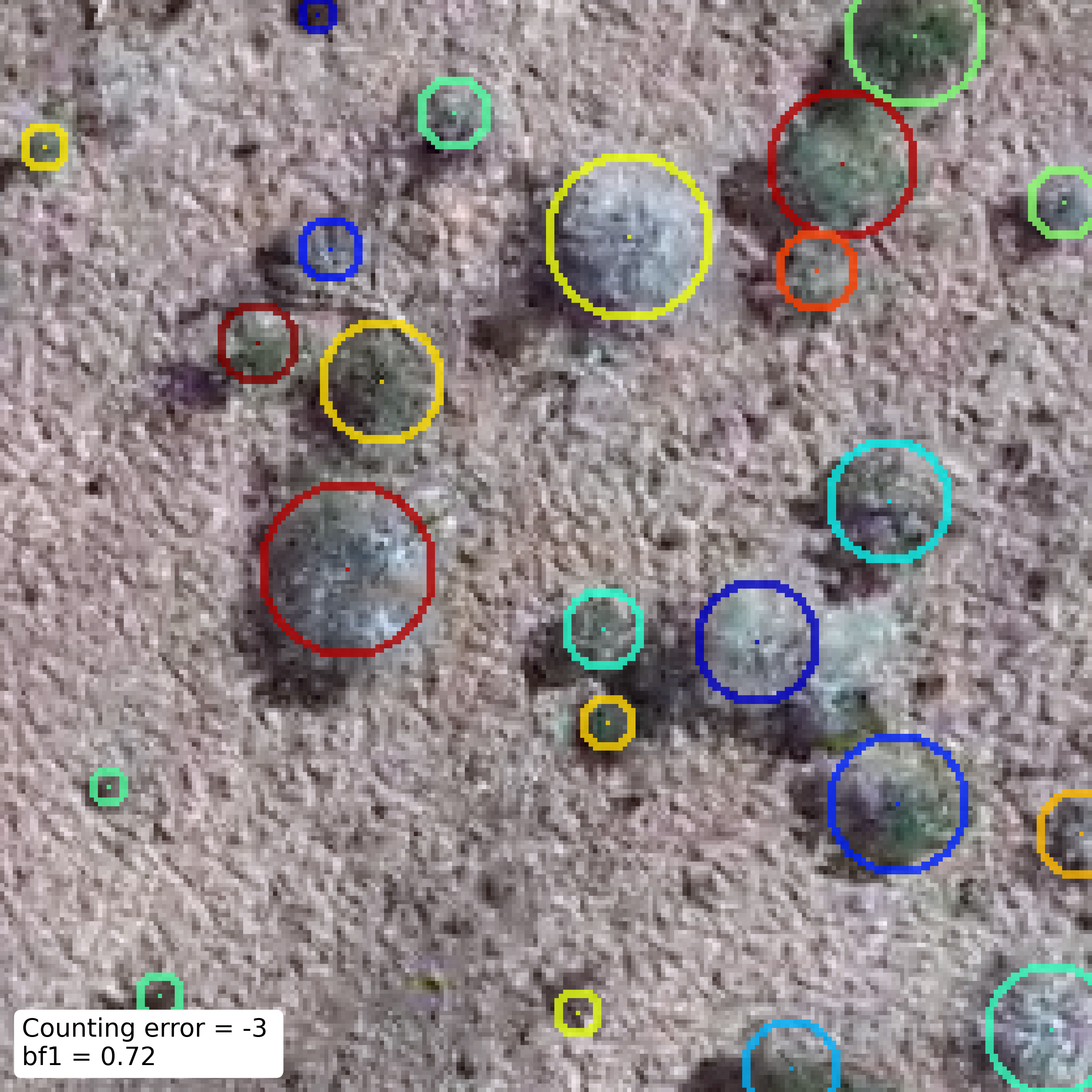}%
}
\hspace{.3ex}
\subfloat[]{\includegraphics[width=.24\linewidth]{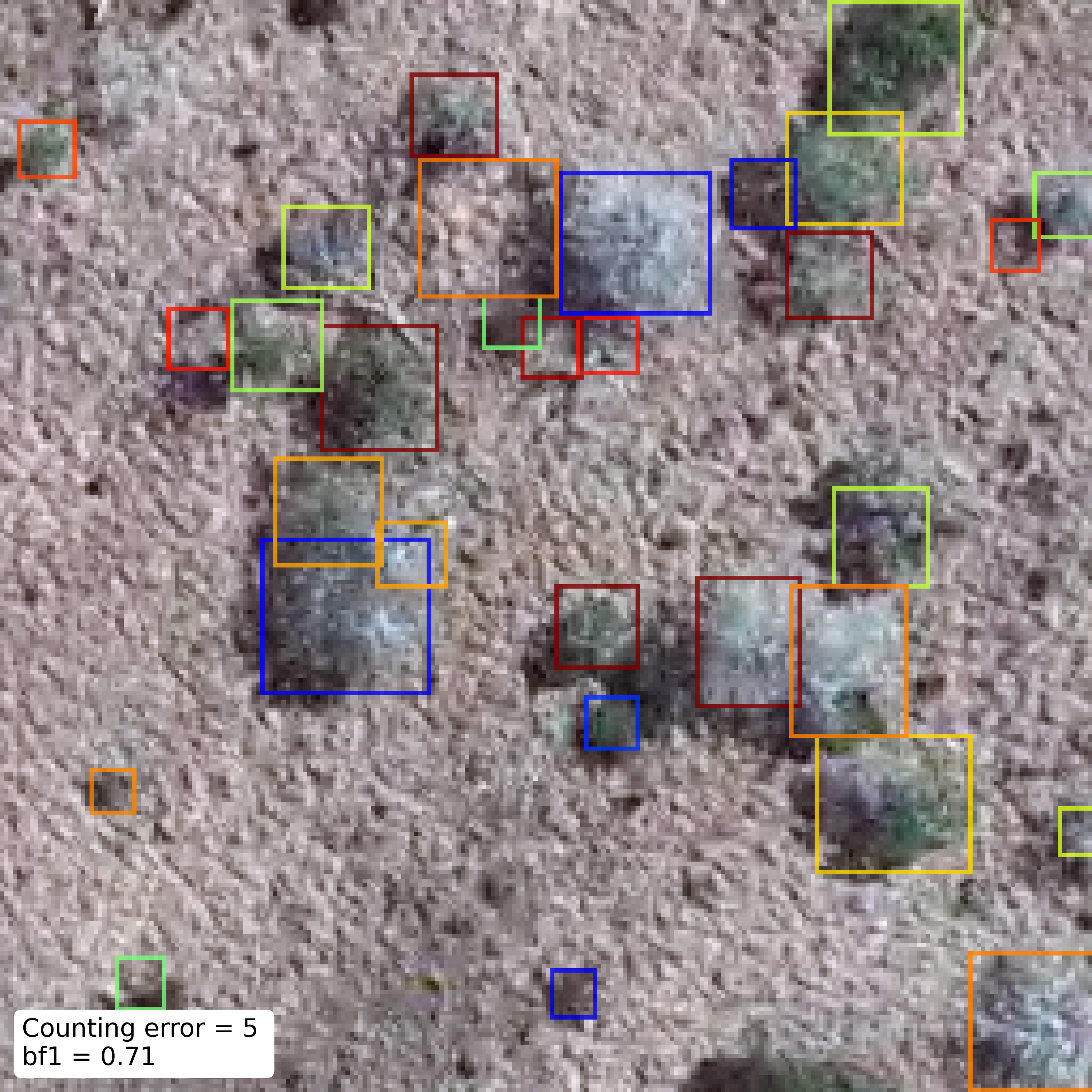}%
}
\caption{Examples of predictions (Rwanda). (a) Input with labels. (b) Segmentation. (c) CenterNet. (d) Heatmap. (e) Point proposal. (f) Box proposal. Colors indicate different instances. bF1 score is indicated, higher is better.}
\label{fig:lowdensityexample}

\end{figure*}

\begin{figure*}[!t]
\captionsetup[subfloat]{captionskip=-1.2em,justification=raggedleft,singlelinecheck=false,font=normalsize,labelfont={color=white, bf}}
\centering
\subfloat[]{\includegraphics[width=.24\linewidth]{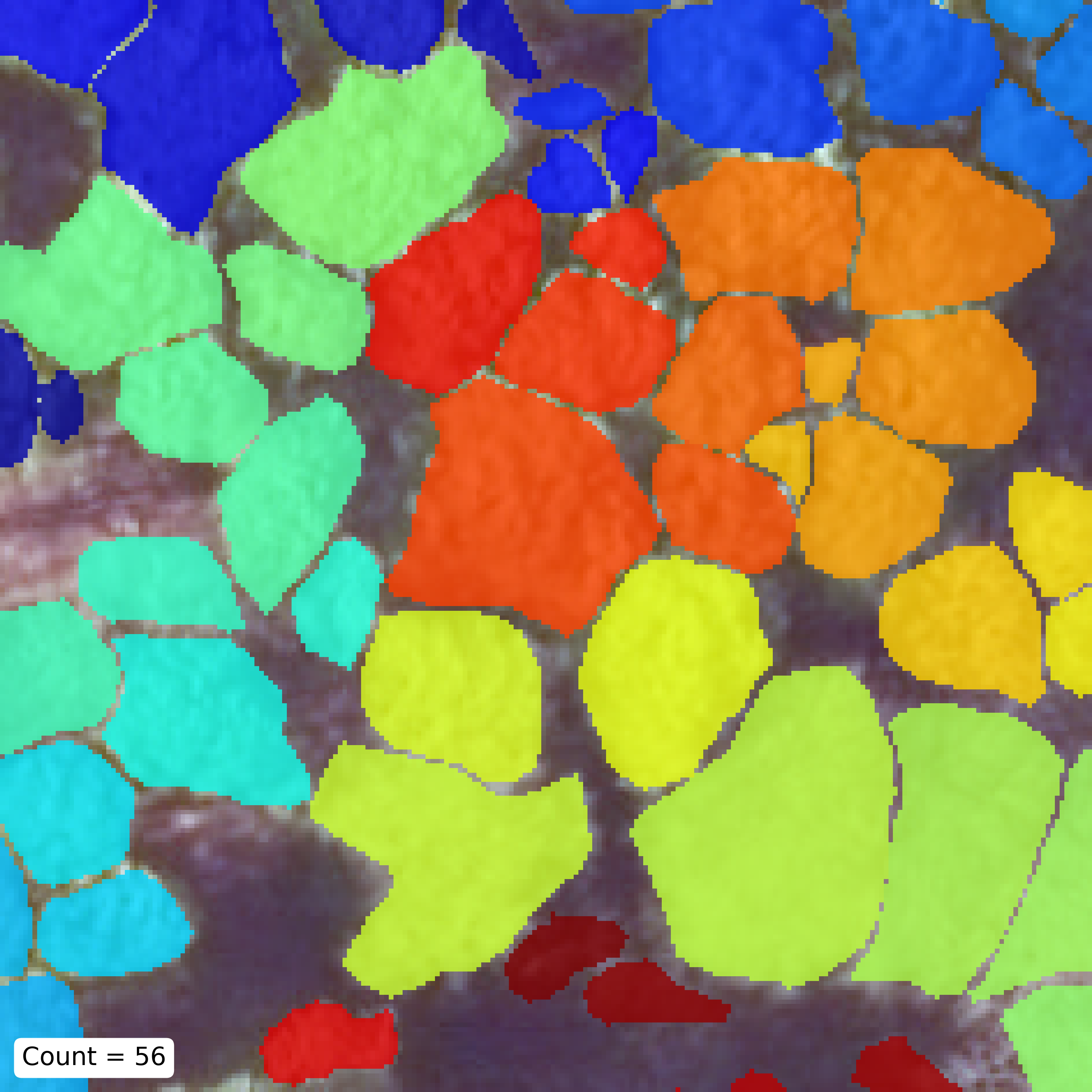}%
% \label{fill}
}
\hspace{.3ex}
\subfloat[]{\includegraphics[width=.24\linewidth]{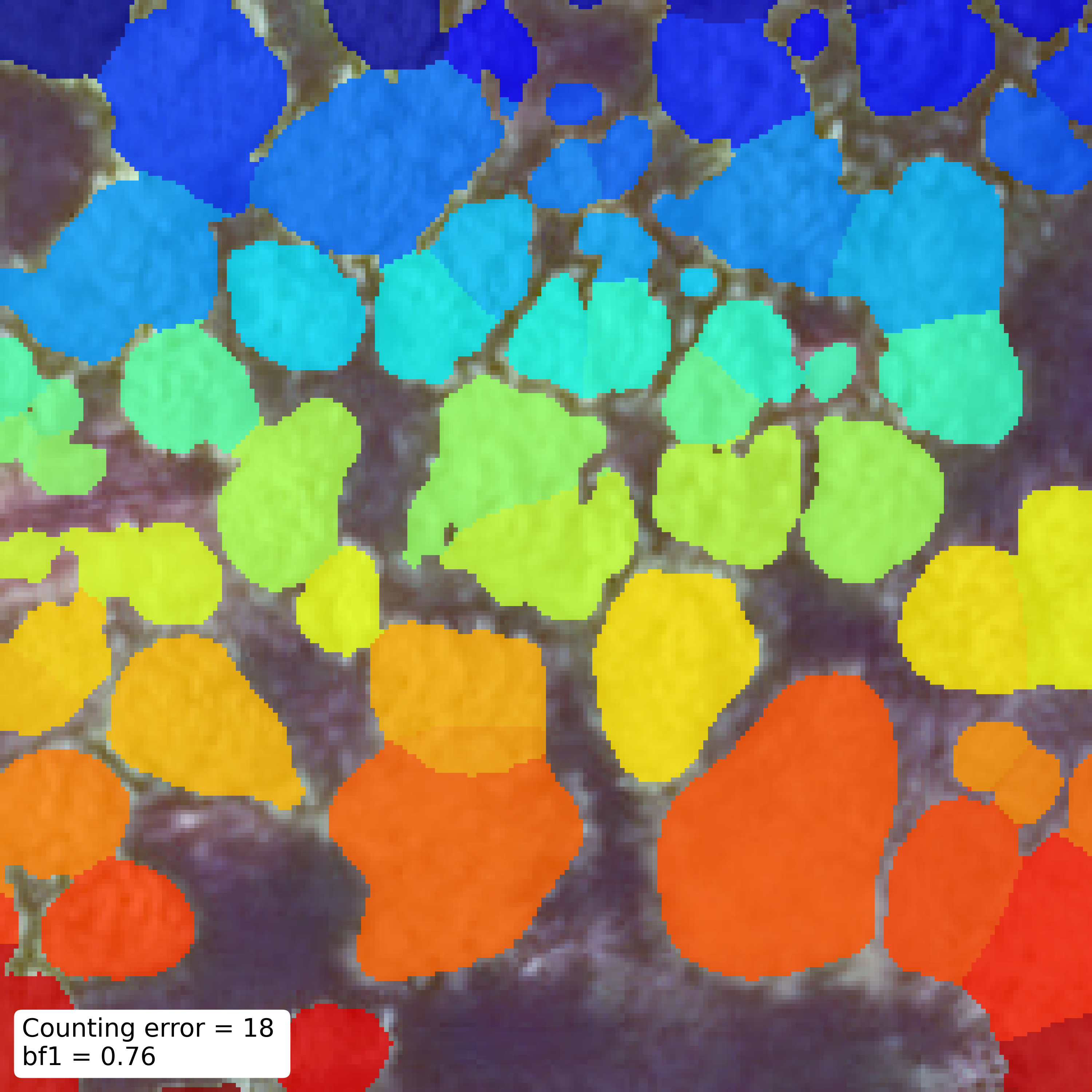}%
% \label{subfig:unet_prob}
}
\hspace{.3ex}
\subfloat[]{\includegraphics[width=.24\linewidth]{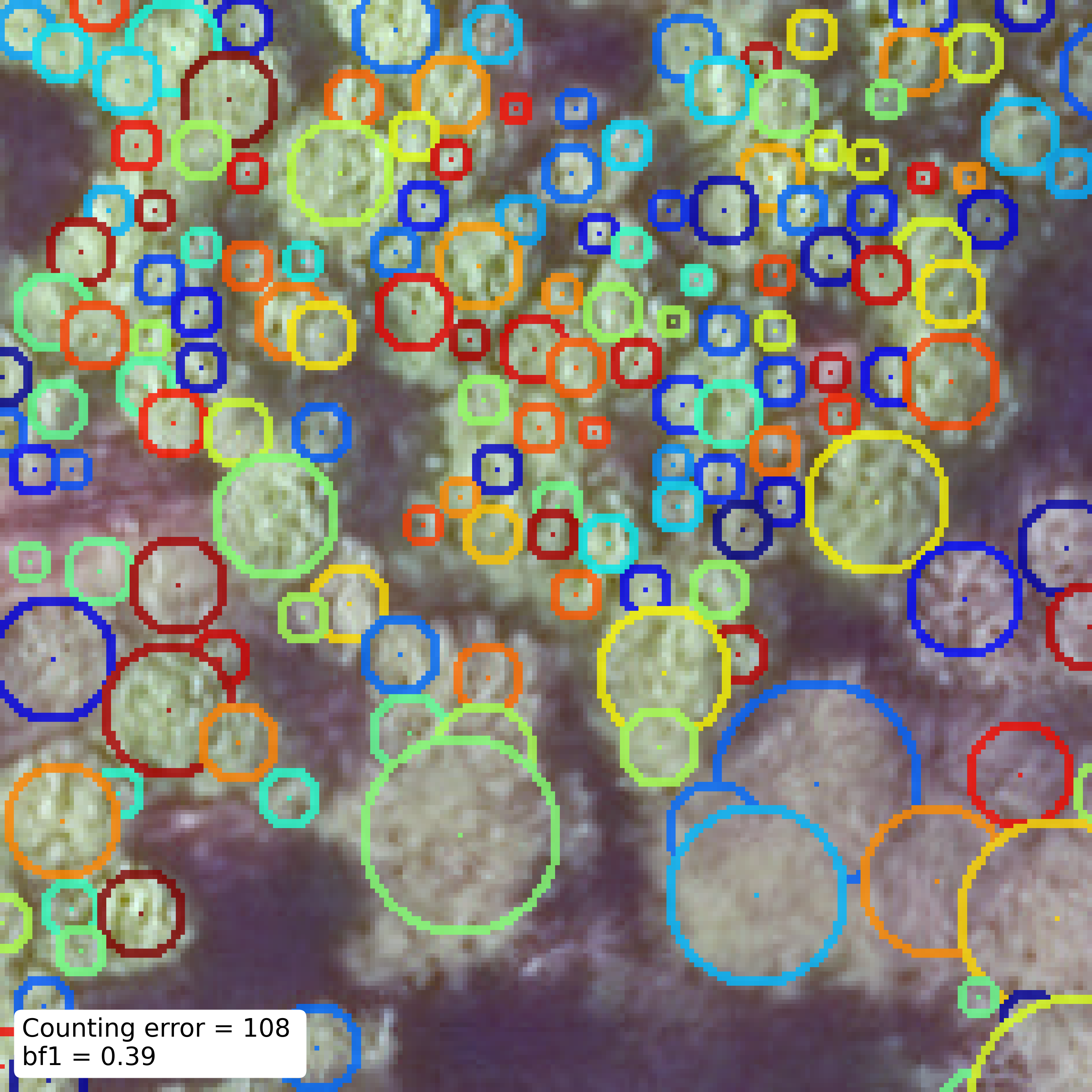}%
% \label{fill2}
}
\\[1ex]
\subfloat[]{\includegraphics[width=.24\linewidth]{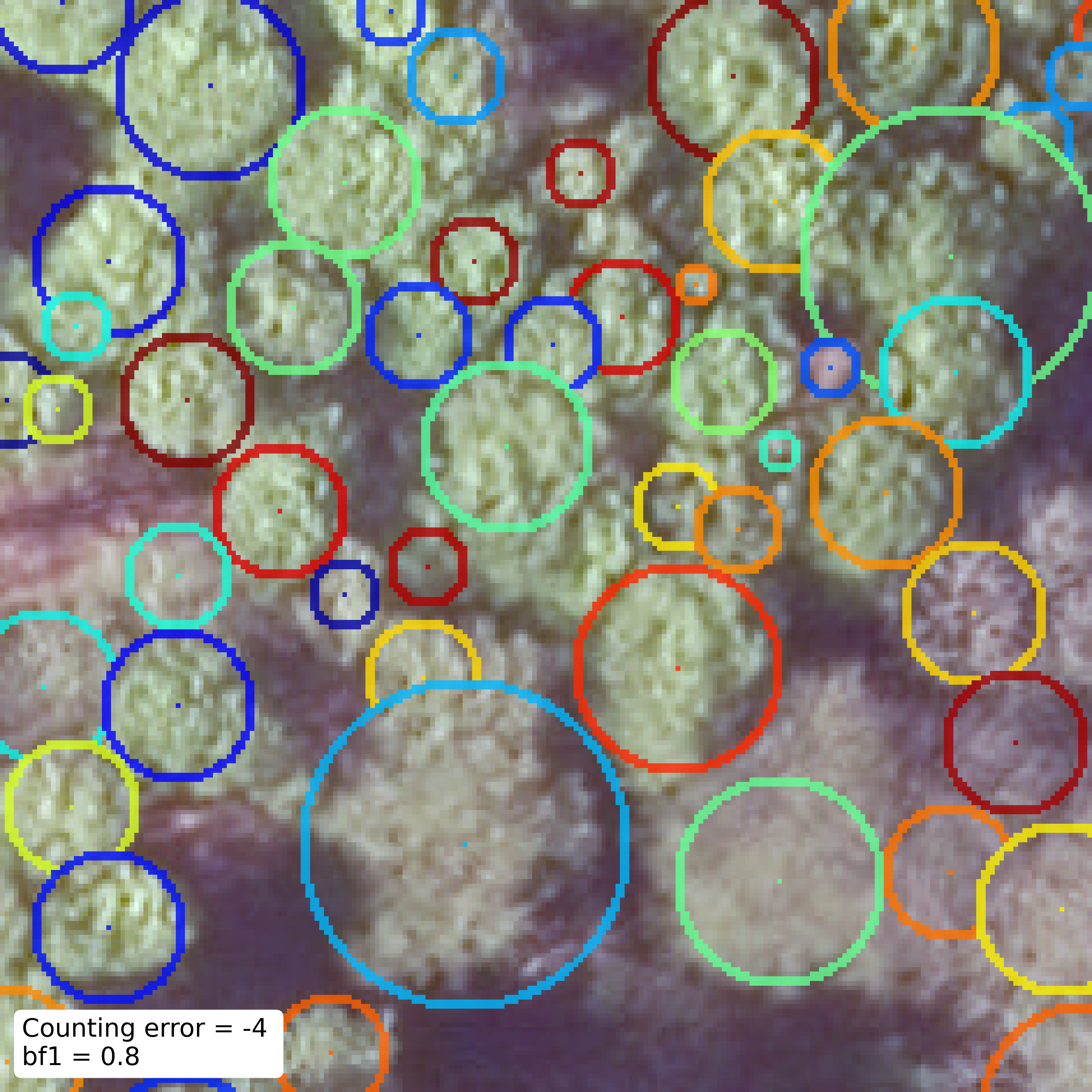}%
% \label{fill3}
}
\hspace{.3ex}
\subfloat[]{\includegraphics[width=.24\linewidth]{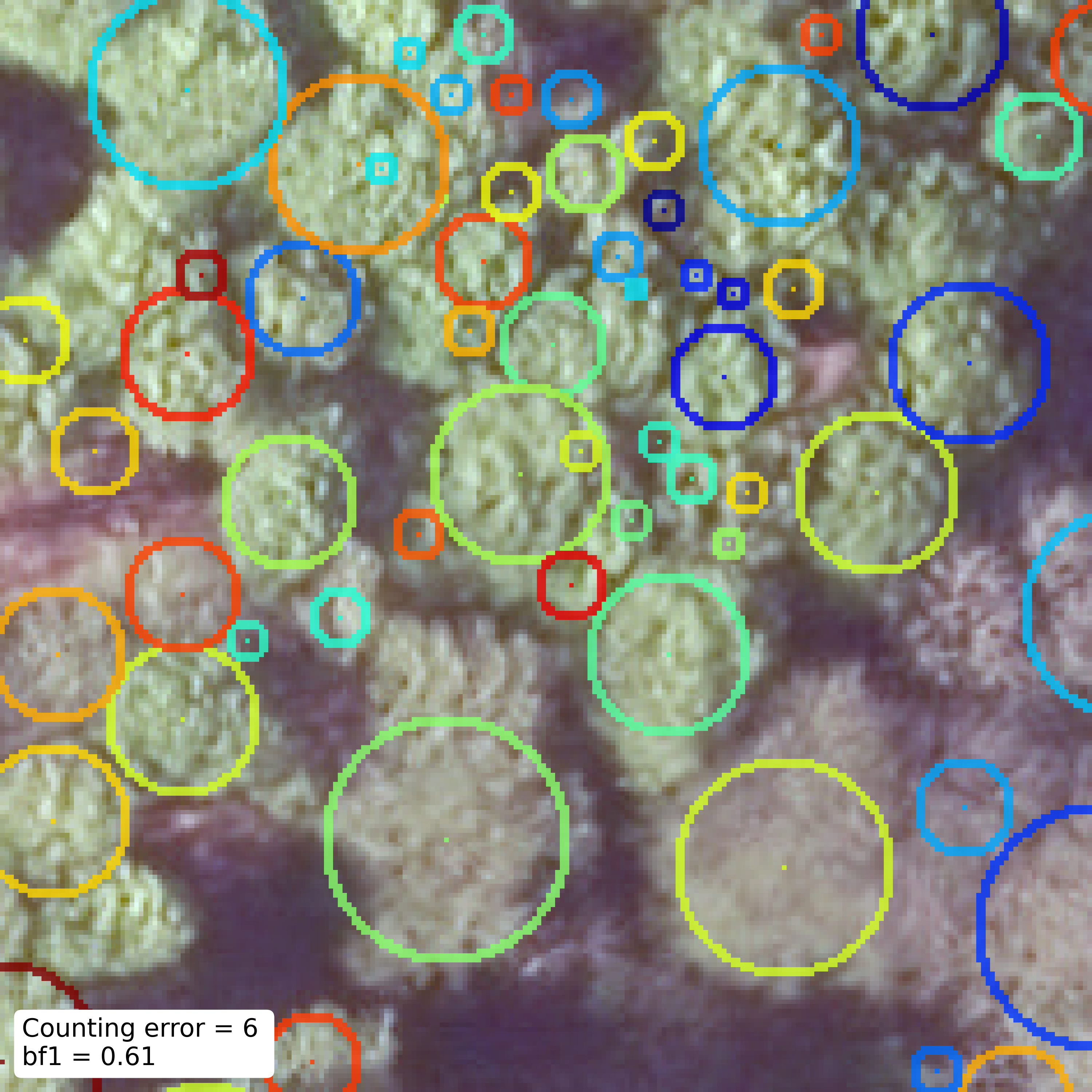}%
% \label{fill4}
}
\hspace{.3ex}
\subfloat[]{\includegraphics[width=.24\linewidth]{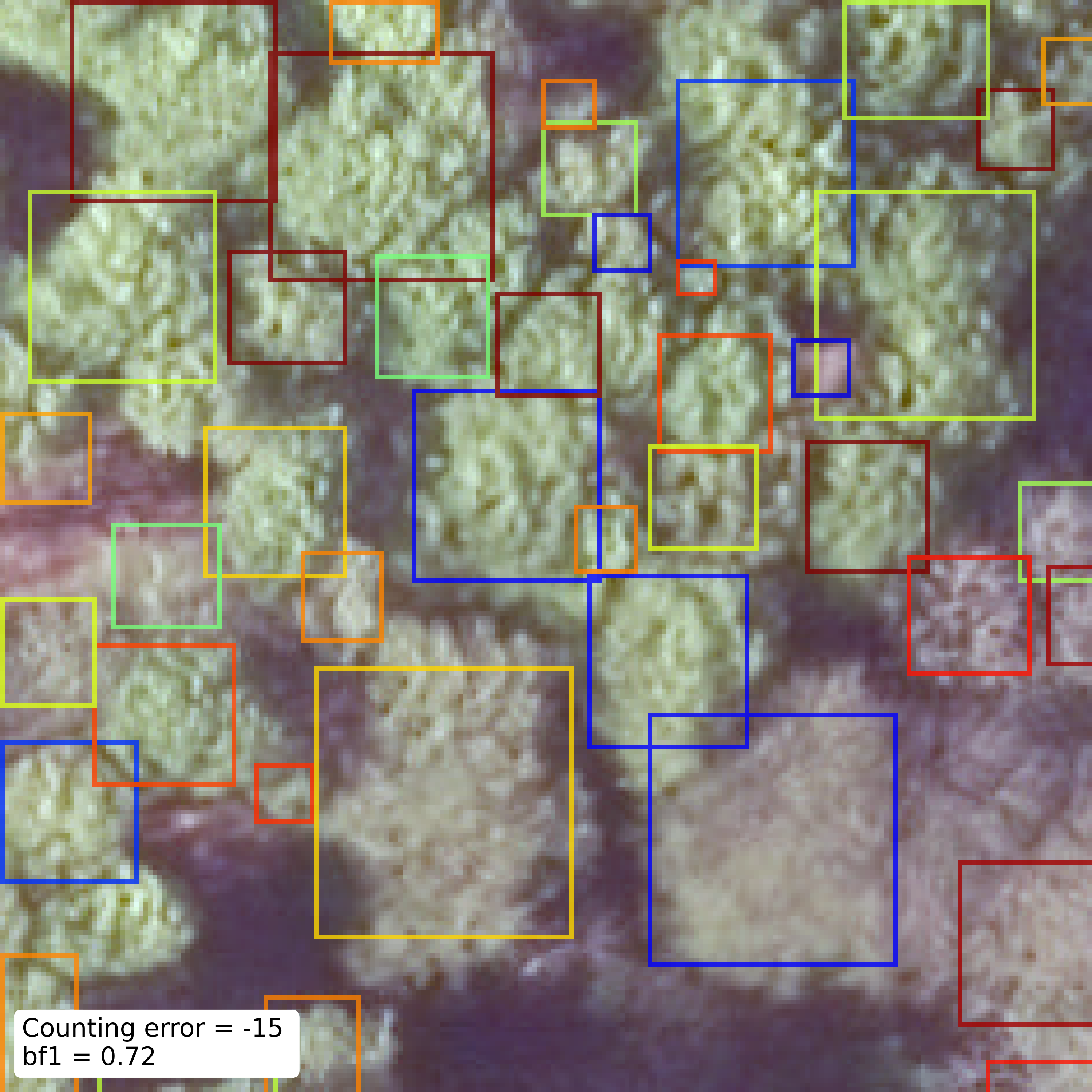}%
% \label{fill5}
}
\caption{Examples of predictions (Denmark). (a) Input with labels. (b) Segmentation. (c) CenterNet. (d) Heatmap. (e) Point proposal. (f) Box proposal. Colors indicate different instances. bF1 score is indicated, higher is better.}
\label{fig:highdensityexample}
\end{figure*}

We conduct a visual inspection to refine our comparison of methods. We compare the 5 methods (with UNet-R50 for heatmap-based methods) on the same train/val split in \Cref{fig:lowdensityexample,fig:highdensityexample}.

The predicted circular approximation drawn on the visual examples is often a good fit for the actual tree cover, and small individual trees are properly separated despite the proximity with neighbors.

The Segmentation approach is visually the most consistent with the actual tree cover. Its limitations come from the splitting and merging effects of the necessary instance separation post-processing step.

\subsection{Computational cost}

It is important to consider the resource usage for scaling up tree mapping to larger areas. We compare frameworks and architectures in Table~\ref{tab:processingtime}, including counting the forward pass time as the average time for computing raw neural network outputs, and the post-processing time as the average time to get individual instances (NMS or instance separation), and the conversion of the outputs to a list of predictions (not necessary for anchor-based approaches). Experiments are done on the same machine with the same train/val split, on a single Nvidia RTX 3090 (24Go VRAM) and AMD 3960X (3.8 GHz, 24 cores) CPU.

The anchor-based approaches are the fastest, with virtually no processing cost, given their efficient architecture and highly optimized NMS. The CenterNet and Heatmap approaches come second with a post-processing time that mostly depends on the number of predictions. UNet is the fastest backbone, with a total time comparable to the anchor-based approaches. The segmentation approach is 5 to 40 times slower due to the necessary process of associating each candidate pixel to an instance.

\begin{table}
\caption{Processing time for 256x256 patches.}
\label{tab:processingtime}
\centering
\begin{adjustbox}{width=\linewidth}
\begin{tabular}{|c|c|c|c|c|c|}
    \hline
    Framework & Architecture & Forward pass time & Post-processing time\\
    \hline
    \multirow{4}{*}{Segmentation} & UNet-R50 & 1.7ms & 81.8ms\\
     & DeepLabV3 & 1.8ms & 92.4 ms\\
     & SegFormer & 3.0ms & 91.3ms\\
     & TransUNet & 2.5ms & 91.9ms\\
    \hline
    \multirow{4}{*}{CenterNet} & UNet-R50 & 1.7ms & 1.4ms\\
     & DeepLabV3 & 1.8ms & 14.3ms\\
     & SegFormer & 3.1ms & 11.3ms\\
     & TransUNet & 2.5ms & 4.0ms\\
    \hline
    \multirow{4}{*}{Heatmap} & UNet-R50 & 1.7ms & 2.1ms\\
     & DeepLabV3 & 1.8ms & 14.9ms\\
     & SegFormer & 3.0ms & 11.0ms\\
     & TransUNet & 2.5ms & 4.7ms\\
    \hline
    Point proposal & P2P & 3.4ms & 0.5ms\\
    \hline
    Box proposal & FasterRCNN & 6.1ms & 0.2ms\\
    \hline
\end{tabular}
\end{adjustbox}
\end{table}

\subsection{Labeling noise}
\label{subsec:labelexp}

We asked two annotators to label a small study set of 13 areas. Annotator A labeled a total of 1527 trees, annotator B 1248. Using distance and size criteria with one-to-many and many-to-one matching, we exhaustively compared annotations to assess the agreement between annotators.

We report in Table~\ref{tab:labelcomparison} the number of situations where trees were not annotated in one set (identity with a connection degree = 0), the number of situations where annotators fully agree, and the number of situations with different levels of merging and splitting. In total, we also noted 6.9\% of N-to-M situations.

\begin{table}
\caption{Differences between labels from two annotators.}
\label{tab:labelcomparison}
\vspace*{5mm}
\centering
\begin{adjustbox}{width=.5\linewidth}
\begin{tabular}{|c|c|c|c|}
    \hline
    Degree & Identity & Split & Merge\\
    \hline
    0 & 35.2\% & - & -\\
    1 & 44.4\% & - & -\\
    2 & - & 5.5\% & 5.6\%\\
    3 & - & 0.7\% & 1.0\%\\
    4 & - & 0.2\% & 0.2\%\\
    5 & - & 0\% & 0.1\%\\
    \hline
\end{tabular}
\end{adjustbox}
\end{table}

Using Bayes' rule and our proposed model of label merging/splitting (see Section~\ref{subsec:labnoise}), we can estimate the posterior probability of having a real tree with a certain crown area, knowing its label. Using Eq~\ref{eq:labelquantity} and assuming that the agreement between two annotators is the same as the agreement between one annotator and the real distribution, we model the probability of having a real crown area given the label crown area as:

\begin{equation}
    p(\mathrm{CA}_l = s_l | \mathrm{CA}_r = s_r) = \begin{cases}
        0.35 & \text{if } s_l = 0 \\
        0.44 & \text{if } s_l = s_r \\
        0.07 \cdot \frac{f_p(q ; \lambda)}{1 - f_p(1 ; \lambda)} & \text{if } s_l = 2s_r, 3s_r, 4s_r \ldots \\
        0.06 \cdot \frac{1 - f_p(1 ; \lambda)}{f_p(q ; \lambda)} & \text{if } s_l = \frac{s_r}{2}, \frac{s_r}{3}, \frac{s_r}{4}, \ldots \\
    \end{cases}
\end{equation}

\begin{figure}[t]
  \centering
   \includegraphics[width=\linewidth]{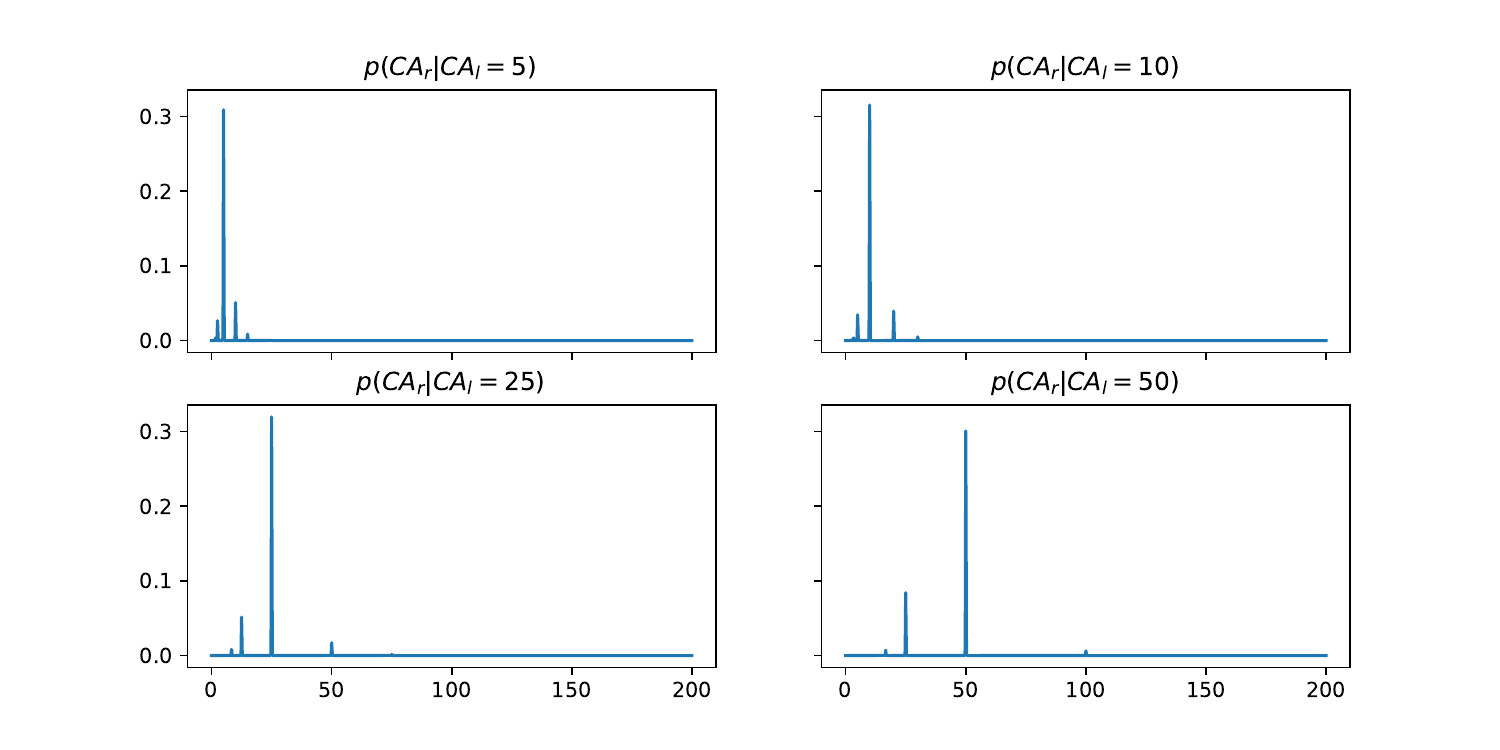}
   \caption{Posterior probability of having real crown area $\mathrm{CA}_r$ depending on labeled crown diameter $\mathrm{CA}_l$.}
   \label{fig:postprobs}
\end{figure}

As an estimation of $p(\mathrm{CA}_r)$, the probability of having a certain real CA, we use the global allometry database \citep{jucker_allometric_2017}, filter on temperate forests and compute the distribution of crown areas from measured crown diameters. We compute $p(\mathrm{CA}_l)$ from the distribution of labeled crown areas. We get a pdf of real CA for each labeled CA, and plot four examples in Figure~\ref{fig:postprobs}. As labeled trees get bigger, there is a higher change that the corresponding real trees have a smaller crown area and were merged in the labeling process. Conversely, smaller labeled trees have a high chance of corresponding to a bigger real tree that was mislabeled by splitting. Note that we simplify the model to a large extent here by considering that splitting/merging multiplies crown areas by an integer, in reality a tree crown could be split (or merged) to (or from) unequally sized crowns.

\begin{figure}[t]
  \centering
   \includegraphics[width=\linewidth]{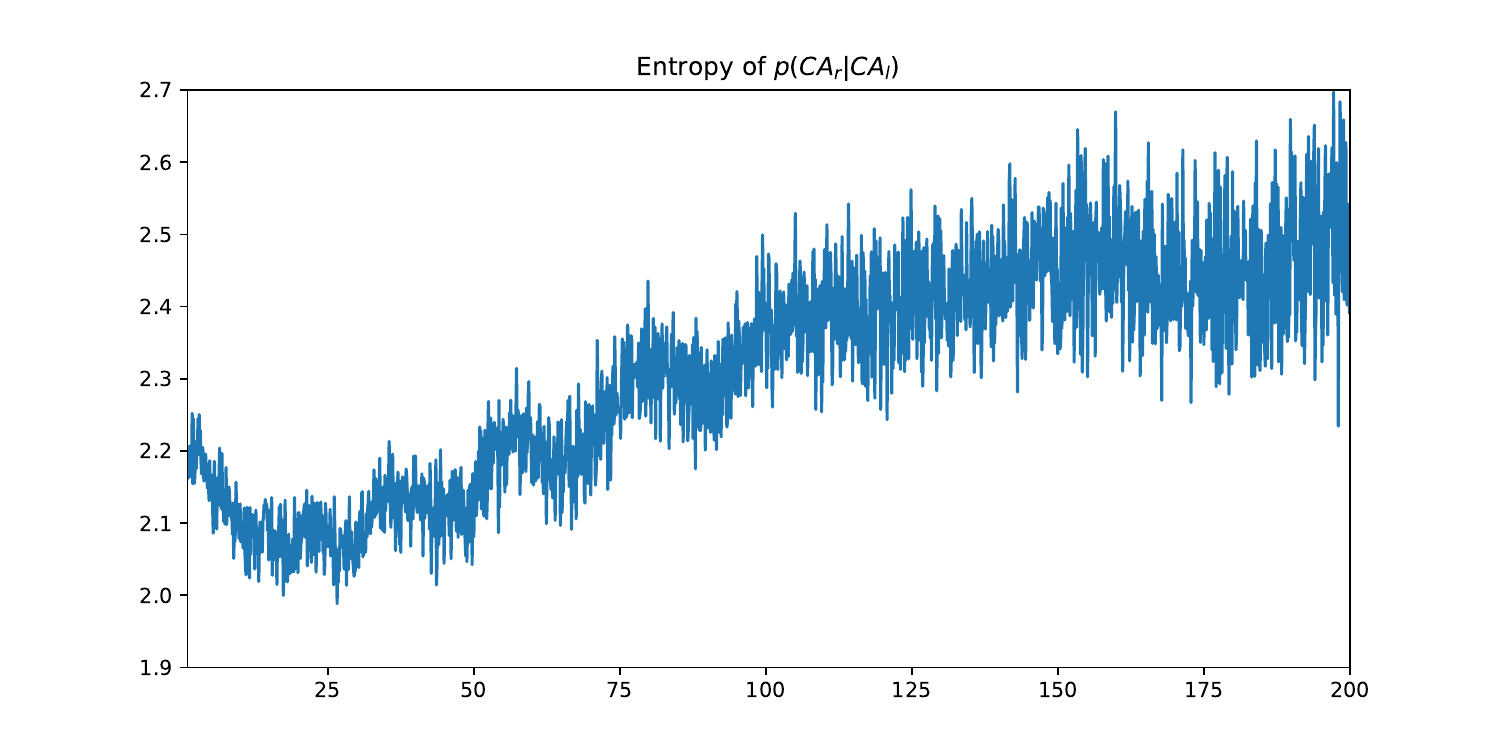}
   \caption{Uncertainty on real crown area depending on the labeled crown area.}
   \label{fig:uncertainty}
\end{figure}

We plot in Figure~\ref{fig:uncertainty} the entropy of $p(\mathrm{CA}_r | \mathrm{CA}_l)$, \ie the uncertainty of the real crown area given a labeled crown area. We note a higher uncertainty on very small trees ($\mathrm{CA} < 10m^2$) and big trees ($\mathrm{CA} > 30m^2$), with a "sweet spot" around $15m^2$ where the uncertainty on the real tree CA is the lowest.

\section{Conclusion}

We introduced an evaluation framework to compare methods and deep learning architectures for mapping of individual trees in remote sensing imagery. By introducing a proxy of N-to-M matching between predicted and labelled trees, we better measure how well methods reproduce the real distribution of trees. Our implementation and comparison of 5 methods and 6 backbone architectures shows different behaviors with different localization thresholds, and provides detailed indicators for picking the optimal method-backbone combination. 

On the two datasets we consider, a recent point-based detection method with integrated one-to-one matching (P2P) achieves the highest detection scores. In terms of architectures, UNet remains overall a strong baseline for high detail mapping, with a performance that matches or surpasses modern Transformer-based architecture without the added complexity. More suprisingly, the Segmentation approach, which relies on precise crown delineation, does not outperform other approaches significantly, indicating that the core of the problem lies in separating trees.

Building on our review of existing methods, we introduced a novel detection approach that relies on a single heatmap to predict location and estimate crown size. Our proposed method shows good detection scores at low localization tolerance, and is particularly good at placing tree centers at the right position. With ensembling, it also achieves the highest detection scores on difficult setups. Notably, we show that the complete shape information commonly labeled with polygons for tree mapping is not necessary to obtain an accurate localization and size estimation of trees, an interesting pathway towards faster and easier annotation that only requires point labels and crown area.

An important aspect of individual tree mapping is labeling. In dense forests, even expert annotators can make errors, which impacts model training and more importantly the evaluation. From an experiment with two annotators we quantified labeling disagreement, and we showed that the matching process has a strong influence on how robust the metrics are to labeling errors. We encourage the community of remote sensing analysts to focus on adding robustness to label noise, or correcting labels, rather than developing new models and architectures.

\section*{Acknowledgments}

DG ackowledge support by the European Union’s Eurostars programme through the C-Trees project, grant number E114613. MB was supported by the European Research Council (ERC) under the European Union’s Horizon 2020 research and innovation programme (grant agreement no. 947757 TOFDRY) and a DFF Sapere Aude grant (no.~9064–00049B). AK, RF, and CI acknowledge support by the Villum Foundation through the project Deep Learning and Remote Sensing for Unlocking Global Ecosystem Resource Dynamics (DeReEco, no.~34306). AK and CI acknowledge support by the Pioneer Centre for AI, DNRF grant number P1. SL was supported by Institut Europlace de Finance. M.M. was supported by a DFF Sapere Aude grant (no. 9064-00049B).

\bibliographystyle{elsarticle-num-names} 
\bibliography{references}
 
%% The Appendices part is started with the command \appendix;
%% appendix sections are then done as normal sections
%% \appendix

\appendix

\section{Segmentation setup}
\label{appx:segmentation}

The segmentation models are trained with binary ground truth segmentation masks. To enforce better instance separation as suggested in \citep{ronneberger_u-net_2015}, we add a precomputed boundary weighting map, used in the loss functions to give higher importance to the areas between instances.

We use a combination of the Focal and Tversky losses, with weighting factors respectively $0.4$ and $0.6$. The Tversky loss has coefficients $0.4$ and $0.6$ for weighting false positives and false negatives respectively.

To separate instances during post-processing, we first identify candidate individuals by computing the Euclidean distance transform of the predicted binary mask and getting the positions of maximum values on a local neighborhood of size $w_{max}$. We keep only one peak position for each local neighborhood. Positive pixels are then exhaustively assigned to the closest candidate. Finally, the resulting instance segmentation map is refined with a majority filter of size $w_{maj}$ to avoid isolated pixels. We set $w_{max} = 15$ and $w_{maj} = 23$ for 256x256 images.

Crown diameter is calculated as the diameter of the disk with the same area as the predicted polygon, similarly to how crown diameter is calculated in the training set for other methods.

\section{CenterNet setup}
\label{appx:hm+prmap}

Target crown diameters for training the models are obtained from the labeled polygons, as the diameter of the disk with the same area as the polygon. 

We use a L1 loss for optimizing the heatmap, and a masked L1 loss for the size map (ignoring values at positions where there is no ground truth). We train the heatmap and size map heads with a 10x base learning rate. During inference, we remove values on the heatmap that are not equal to the max value in a 11x11 sliding window, before identifying centers as positions with values above a threshold set at $0.5$.

\section{Heatmap setup}
\label{appx:hm}

We train models with a L1 loss. The target heatmaps are generated by processing the segmentation masks. Individual trees are separated with the labeled masks,and we identify the centers and calculate the crown diameter as the diameter of the disk that would have the same area. We subsequently draw the heatmaps by generating a Gaussian kernel of standard deviation depending on the crown diameter, and normalize to have a value of 1 at the center. To enforce better separation between instances, we only keep values within the original polygon boundaries.

For inference, we first perform non-maximum suppression with a simple max pooling operation: values that are not equal to the max value in a 11x11 sliding window are ignored. We identify centers at locations with a value $> 0.6$. At each center, we compute the zero-normalized cross-correlation between the 25x25 patch around it and the Gaussian filters. The filter with the highest cross-correlation gives an estimate of the size of the tree at this location. We found that using 48 filters yields a good compromise between size estimation and computation time. $\sigma$ values are ranging on a logarithmic scale from $0.3$ to $25.0$.

\section{Box detection setup}
\label{appx:boxdetection}

Box detection approaches have a tendency to over-predict, making non-maximum suppression (NMS) an essential step. As in the original paper of FasterRCNN, we implement this by keeping the box with the highest class score when multiple boxes overlap. In our case however, we can make use of the fact that trees generally don't overlap significantly and set the IOU threshold low, here at $0.01$. We verified this assumption on a validation split and confirm that this gives the highest detection rate at $\gamma = 2$.

\section{Point detection setup}
\label{appx:pointdetection}

The Hungarian algorithm cannot be applied to an empty ground truth (\ie a patch without any tree). We consequently trained the models with the point proposal loss on the fraction of training data that contains trees. 
The point detection framework, similarly to the box detection model, requires NMS to avoid redundant predictions. We use the same box-based NMS as FasterRCNN, to maintain consistency with the original method. However, we note that our main conclusion about the circular approximation of trees opens the way for a more suited disk-based NMS.

\section{Feature extractors}
\label{appx:fe}

We consider the following backbones:

\begin{enumerate}
    \item UNet \citep{ronneberger_u-net_2015} is a very popular architecture for segmentation. Its novelty is a U-shaped architecture with an encoder and a decoder, and the intermediate features being linked by skip connections allowing high geometric consistency between input and output. This is particularly important for approaches relying on pixel-wise losses between the prediction and the annotation, such as segmentation (with classes) or heatmap-based detection (with presence probability). Here we use a variant with a ResNet50-based \citep{he_deep_2016} encoder.
    \item DeepLabV3 \citep{chen_rethinking_2017} is an example of a model which does not use the encoder-decoder structure as UNet, but instead applies a single convolutional pipeline. Using pyramids of atrous convolutions (convolutions with a dilated kernel), the model obtains an understanding of context while maintaining local accuracy. A downside of this approach is that the output resolution must be kept relatively low due to memory limitations, implying a bilinear interpolation at the output to get the same size as the input (necessary for the segmentation and heatmap-based approaches).
    \item SegFormer \citep{xie_segformer_2021} is a convenient baseline for pure transformer-based feature extraction. Transformer architectures, being relatively recent, are hard to reuse due to inefficient processing or task-specific design choices (the 16 × 16 blocks of ViT \citep{dosovitskiy_image_2021} for example). SegFormer solves some of these problems with a hierarchical architecture and an optimized attention block.
    \item TransUNet \citep{chen_transunet_2021}, a revisited version of UNet with a transformer-based encoder, aims at proposing the best of both worlds: the long-range modeling of transformers with the geometric consistency of convolutions. 
\end{enumerate}

\begin{table*}
\caption{Backbone architectures}
\label{tab:backbones}
\centering
\begin{adjustbox}{width=0.7\linewidth}
\begin{tabular}{|c|c|c|c|c|c|}
    \hline
    Name & Type & Architecture & \# parameters & \# Anchors & Output downscaling \\
    \hline
    \multicolumn{6}{|c|}{Anchor-based} \\
    \hline
    P2P-VGG16 & Conv. & Direct & 24M & 1024 & - \\
    FasterRCNN-R50 & Conv. & Direct & 41M & 16368 & - \\
    \hline
    \multicolumn{6}{|c|}{Heatmap-based} \\
    \hline
    UNet-R50 & Conv. & Encoder-Decoder & 33M & - & x1  \\
    DeeplabV3 & Conv. & Direct & 40M & - & x8 \\
    SegFormer & Transf. & Encoder-Decoder & 45M & -  & x4 \\
    TransUNet & Conv. + Transf. & Encoder-Decoder & 105M & - & x1\\
    \hline
\end{tabular}
\end{adjustbox}
\end{table*}

\section{One-to-one matching evaluation}

For comparison, we provide the benchmarking results with one-to-one matching in Table~\ref{tab:11detresults}. The Segmentation approach performs better on Rwanda in this case, and P2P remains the most accurate on Denmark. 

\begin{table*}
\caption{Tree detection benchmarking results with one-to-one matching.}
\label{tab:11detresults}
\centering
  \begin{adjustbox}{width=\textwidth}
  \begin{tabular}{|cc|ccc|ccc|ccc|}
    \hline
    \multirow{2}{*}{Framework} & \multirow{2}{*}{Architecture} & \multicolumn{3}{c|}{$\gamma = 0.5$} & \multicolumn{3}{|c|}{$\gamma = 1$} & \multicolumn{3}{|c|}{$\gamma = 2$}\\
     & & Precision $\uparrow$ & Recall $\uparrow$ & $\mathrm{F_1}$ $\uparrow$ & Precision $\uparrow$ & Recall $\uparrow$ & $\mathrm{F_1}$ $\uparrow$ & Precision $\uparrow$ & Recall $\uparrow$ & $\mathrm{F_1}$ $\uparrow$ \\
    \hline
    \multicolumn{11}{|c|}{\large Denmark} \\
    \hline
    \multirow{1}{*}{Segmentation} & UNet-R50 & $33.6\scriptstyle{\pm1.1}$ & $33.0\scriptstyle{\pm3.1}$ & $32.1\scriptstyle{\pm2.0}$ & $47.0\scriptstyle{\pm1.9}$ & $45.8\scriptstyle{\pm5.7}$ & $44.2\scriptstyle{\pm3.6}$ & $54.5\scriptstyle{\pm1.4}$ & $52.8\scriptstyle{\pm5.4}$ & $51.0\scriptstyle{\pm3.3}$
\\
    \hline
    \multirow{1}{*}{CenterNet} & UNet-R50 & $30.6\scriptstyle{\pm2.5}$ & $34.0\scriptstyle{\pm5.7}$ & $30.1\scriptstyle{\pm3.0}$ & $43.2\scriptstyle{\pm4.4}$ & $46.2\scriptstyle{\pm7.1}$ & $41.2\scriptstyle{\pm3.6}$ & $50.6\scriptstyle{\pm5.5}$ & $53.0\scriptstyle{\pm7.9}$ & $47.4\scriptstyle{\pm3.9}$
\\
    \hline
    \multirow{1}{*}{Heatmap} & UNet-R50 & $36.9\scriptstyle{\pm1.8}$ & $33.2\scriptstyle{\pm4.5}$ & $33.1\scriptstyle{\pm3.6}$ & $48.5\scriptstyle{\pm0.8}$ & $42.4\scriptstyle{\pm5.2}$ & $42.5\scriptstyle{\pm3.5}$ & $54.1\scriptstyle{\pm1.8}$ & $47.1\scriptstyle{\pm6.2}$ & $47.1\scriptstyle{\pm4.2}$
\\
     \hline
    \multirow{1}{*}{Point proposal} & P2P & $34.5\scriptstyle{\pm2.3}$ & $29.5\scriptstyle{\pm1.7}$ & $30.8\scriptstyle{\pm1.3}$ & $54.6\scriptstyle{\pm4.0}$ & $46.0\scriptstyle{\pm2.0}$ & $48.0\scriptstyle{\pm1.4}$ & $64.9\scriptstyle{\pm4.4}$ & $55.0\scriptstyle{\pm2.6}$ & $57.0\scriptstyle{\pm1.4}$
\\
    \hline
    \multirow{1}{*}{Box proposal} & FasterRCNN & $35.9\scriptstyle{\pm1.4}$ & $36.4\scriptstyle{\pm0.8}$ & $33.8\scriptstyle{\pm0.8}$ & $49.6\scriptstyle{\pm2.3}$ & $52.6\scriptstyle{\pm2.2}$ & $47.1\scriptstyle{\pm0.9}$ & $56.0\scriptstyle{\pm2.9}$ & $60.3\scriptstyle{\pm2.5}$ & $53.3\scriptstyle{\pm1.2}$

 \\
    \hline
    \multicolumn{11}{|c|}{\large Rwanda} \\
    \hline
    \multirow{1}{*}{Segmentation} & UNet-R50 & $16.2\scriptstyle{\pm0.5}$ & $19.8\scriptstyle{\pm1.0}$ & $16.1\scriptstyle{\pm0.7}$ & $36.9\scriptstyle{\pm0.8}$ & $47.2\scriptstyle{\pm1.6}$ & $37.4\scriptstyle{\pm1.1}$ & $50.7\scriptstyle{\pm1.1}$ & $66.5\scriptstyle{\pm1.6}$ & $52.0\scriptstyle{\pm1.2}$\\
    \hline
    \multirow{1}{*}{CenterNet} & UNet-R50 & $15.1\scriptstyle{\pm0.6}$ & $17.6\scriptstyle{\pm1.7}$ & $14.2\scriptstyle{\pm0.6}$ & $35.4\scriptstyle{\pm0.7}$ & $42.5\scriptstyle{\pm4.2}$ & $33.5\scriptstyle{\pm1.7}$ & $49.0\scriptstyle{\pm1.0}$ & $60.0\scriptstyle{\pm6.4}$ & $46.8\scriptstyle{\pm2.7}$\\
    \hline
    \multirow{1}{*}{Heatmap} & UNet-R50 & $15.0\scriptstyle{\pm1.0}$ & $19.6\scriptstyle{\pm1.0}$ & $15.1\scriptstyle{\pm0.9}$ & $34.9\scriptstyle{\pm2.1}$ & $45.9\scriptstyle{\pm2.3}$ & $34.9\scriptstyle{\pm2.8}$ & $47.8\scriptstyle{\pm3.0}$ & $64.0\scriptstyle{\pm4.1}$ & $48.2\scriptstyle{\pm4.4}$\\
     \hline
    \multirow{1}{*}{Point proposal} & P2P & $14.0\scriptstyle{\pm0.9}$ & $17.7\scriptstyle{\pm0.9}$ & $14.2\scriptstyle{\pm0.8}$ & $33.7\scriptstyle{\pm1.6}$ & $44.3\scriptstyle{\pm2.0}$ & $34.4\scriptstyle{\pm1.3}$ & $48.8\scriptstyle{\pm2.4}$ & $66.4\scriptstyle{\pm3.9}$ & $50.4\scriptstyle{\pm1.9}$\\
    \hline
    \multirow{1}{*}{Box proposal} & FasterRCNN & $15.1\scriptstyle{\pm0.2}$ & $20.6\scriptstyle{\pm1.2}$ & $15.1\scriptstyle{\pm0.3}$ & $33.5\scriptstyle{\pm0.9}$ & $47.4\scriptstyle{\pm1.3}$ & $34.3\scriptstyle{\pm0.3}$ & $46.0\scriptstyle{\pm2.0}$ & $67.6\scriptstyle{\pm0.9}$ & $47.9\scriptstyle{\pm1.5}$\\
    \hline
    
\end{tabular}
\end{adjustbox}
\end{table*}

\section{Label noise model}
\label{appx:labelnoisemodel}

\begin{figure}[t]
  \centering
   \includegraphics[width=\linewidth]{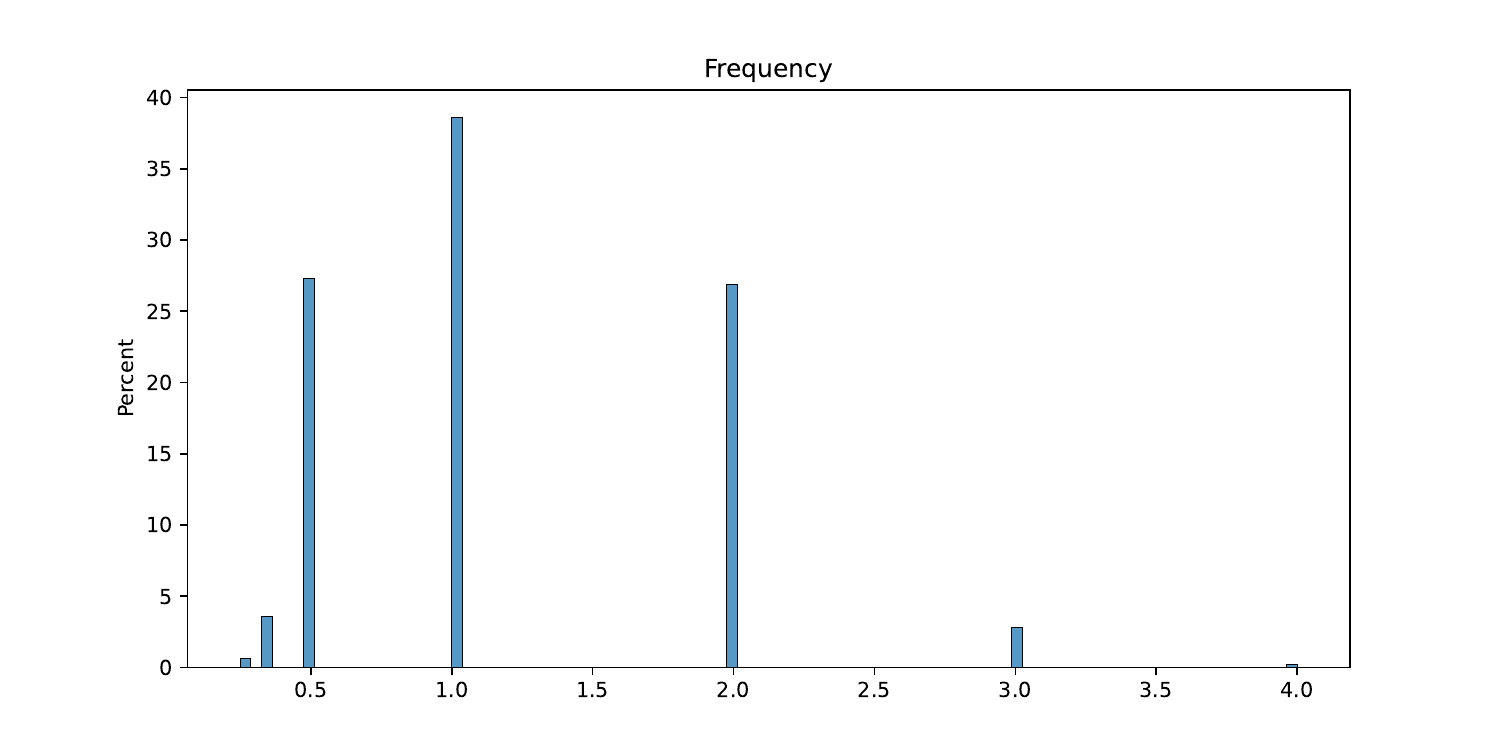}
   \caption{Proposed model of labeling errors as a mixture of Poisson distributions, with $p_1^l = 0.4$ and $\lambda = 0.25$.}
   \label{fig:labelnoisemodelhist}
\end{figure}

We plot in Figure~\ref{fig:labelnoisemodelhist} our proposed pdf of labeling errors (Eq. \ref{eq:labelquantity}). 

%% \section{}
%% \label{}

%% For citations use: 
%%       \citet{<label>} ==> Jones et al. [21]
%%       \citep{<label>} ==> [21]
%%

%% If you have bibdatabase file and want bibtex to generate the
%% bibitems, please use
%%

\end{document}